\documentclass[10pt,twocolumn,letterpaper]{article}

\usepackage[pagenumbers]{cvpr} 

\usepackage{enumitem}
\usepackage{graphicx}
\graphicspath{{figures/}}
\usepackage{subcaption}
\usepackage{multirow}
\usepackage{placeins}
\usepackage{balance}
\usepackage{algorithm}      
\usepackage{algpseudocode}  
\usepackage{array}
\usepackage{cuted}    
\usepackage{capt-of}
\usepackage[normalem]{ulem}

\usepackage{dsfont} 

\usepackage[accsupp]{axessibility}

\definecolor{cvprblue}{rgb}{0.21,0.49,0.74}
\usepackage[pagebackref,breaklinks,colorlinks,allcolors=cvprblue]{hyperref}

\def\paperID{44872} 
\def\confName{CVPR}
\def\confYear{2026}

\title{Dynamic Full-body Motion Agent with Object Interaction \\via Blending Pre-trained Modular Controllers}
\author{
Sanghyeok Nam$^*$ \hspace{0.8em}
Byoungjun Kim$^*$ \hspace{0.8em}
Daehyung Park \hspace{0.8em}
Tae-Kyun Kim\\
KAIST\\
{\tt\small \{sang990701, braian98, daehyung, kimtaekyun\}@kaist.ac.kr}\\
{\small $^*$Equal contribution}\\
\vspace{1cm}
{\small Project Page: \href{https://yurangja99.github.io/dynamic-hoi/}{\texttt{https://yurangja99.github.io/dynamic-hoi/}}}
\vspace{-1.5cm}
}

\begin{document}
\maketitle
\begin{strip}
  \centering
  \vspace{-1cm}
  \includegraphics[width=\textwidth]{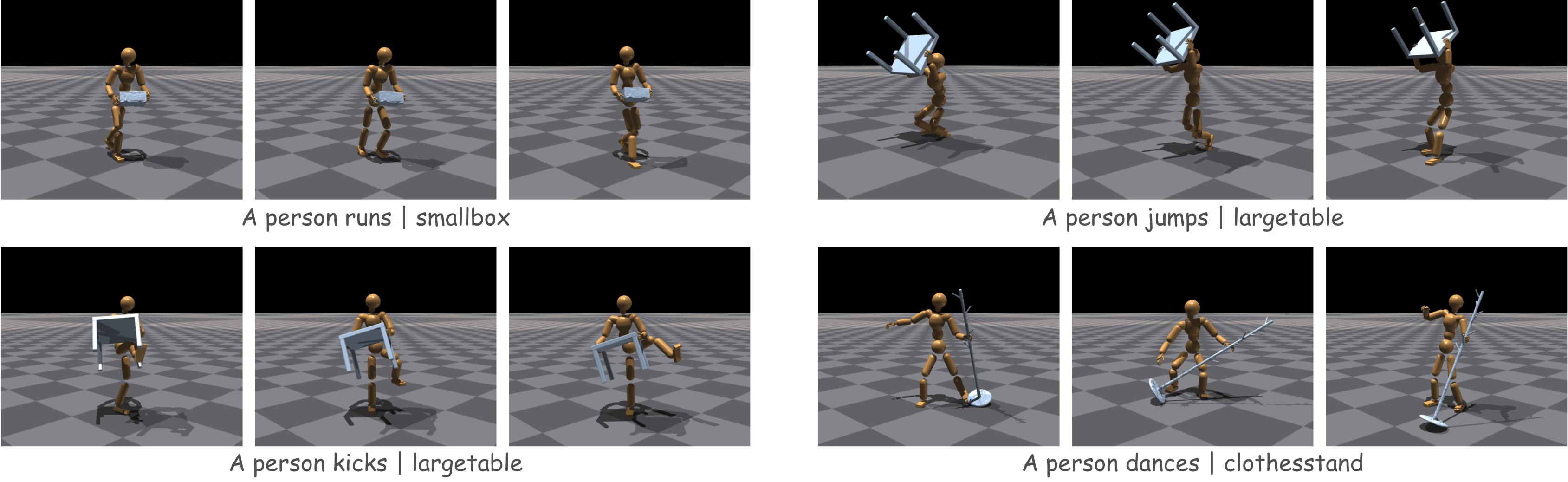}
  \captionsetup{hypcap=false}
  \captionof{figure}{
  We propose a novel framework that blends pretrained experts from distinct motion domains, enabling dynamic and contact-rich human-object interaction (HOI) generation and imitation in physics simulation. Across diverse motion styles and HOI categories, the blended controller demonstrates versatility and robustness while maintaining training efficiency—suggesting a scalable path toward agent training using pretrained modular controllers.}
  \label{fig:teaser}
  \vspace{-1mm}
\end{strip}
\begin{abstract}
Generating physically plausible dynamic motions of human-object interaction (HOI) remains challenging, mainly due to existing HOI datasets limited to static interactions, and pretrained agents capable of either dynamic full-body motions without objects or static HOI motions. Recent works such as InsActor and CLoSD generate HOI motions in planning and execution stages, are yet limited to either static or short-term contacts e.g. striking. In this work, we propose a framework that fulfills dynamic and long-term interaction motions such as running while holding a table, by combining pretrained motion priors and imitation agents in planning and execution stages. In the planning stage, we augment HOI datasets with dynamic priors from a pretrained human motion diffusion model, followed by object trajectory generation. This plans dynamic HOI sequences. In the execution stage, a composer network blends actions of pretrained imitation agents specialized either for dynamic human motions or static HOI motions, enabling spatio-temporal composition of their complementary skills. Our method over relevant prior-arts consistently improves success rates while maintaining interaction for dynamic HOI tasks. Furthermore, blending pretrained experts with our composer achieves competitive performance in significantly reduced training time. Ablation studies validate the effectiveness of our augmentation and composer blending. 
\end{abstract}    
\section{Introduction}
\label{sec:intro}







Real-world humanoid robots are asked to perform dynamic Human-Object Interactions (HOI) motions across diverse, goal-directed scenarios. However, training high-DoF humanoids requires a large amount of high-quality motion data. Yet existing HOI datasets are dominated by static, short-horizon interactions in a narrow motion domain. 
Collecting a scaled dynamic HOI dataset is effort-demanding, a recent work DAViD~\cite{david} proposes synthesizing dynamic HOI motions. It adapts a pretrained motion diffusion model~\cite{tevet2023human} to a HOI domain via LoRA~\cite{hu2022lora}. This, however, requires training category-specific LoRA adapters for individual object classes, and the generated motions do not fulfill contact consistency.

Meanwhile, training a high-DoF agent from scratch remains slow and prone to overfitting. Recent studies attempt to orchestrate various pretrained models including Large Language Models (LLMs), on top of RL, sim2real techniques, to demonstrate and learn the kinematically human-like robots. Vision–Language–Action (VLA) models map egocentric visual observations and natural-language instructions to robot actions via vision encoders with an LLM-based control head \cite{pmlr-v229-zitkovich23a,bu2025univla,deng2025graspvla,figure2025helix,xu2024humanvla,ding2025humanoid}. Despite prevailing paradigms,
object-aware contact reasoning is rarely integrated with high-dynamics, full-body control. Humanoid instantiations often have a low-DoF, are under quasi-static and constrained by specific hand-object contact geometry, limiting applicability to dynamic, contact-rich HOI motions.
Contacts are often approximated or deferred to heuristics, while goal pursuit and agility are optimized in isolation, yielding motions brittle once executed in physical world. 
Recent works such as InsActor \cite{ren2023insactor} and CLoSD \cite{tevet2025closd} generate HOI motions in planning and execution stages to generate diverse and dynamic motions. They adopt a diffusion-based text-to-motion model for planning, then RL imitation learning. They, however, simply formulate HOI tasks to target reaching problems, thus neglecting grasp/contact modeling (e.g., assuming massless objects or overemphasizing collision avoidance) and exhibiting plan–execution discrepancy when planned contacts are physically infeasible.

To tackle the aforementioned limitations, we propose a novel two-stage HOI framework that couples contact-consistent planning with agent compositional execution. In the planning stage, we synthesize diverse, text-conditioned HOI plans by steering an AMASS-pretrained MDM~\cite{tevet2023human, Mahmood_2019_ICCV} with HOI motion priors from FullBodyManip~\cite{li2023object} via interaction-consistent inpainting. Conditioned by the human motion plan, we recover object trajectories by timestep-wise rigid alignment, yielding dynamic human motions with coherent hand–object contacts. In the execution stage, a composer network is learnt on the synthetic motions from the planning stage, by spatio-temporal gating two pretrained imitation agents—InterMimic for HOI \cite{xu2025intermimic} and PHC for full-body agility \cite{Luo2023PerpetualHC}—thereby embodying complementary skills to physical action space. By learning joint/time-dependent mixture weights, the composer allocates responsibility across the kinematic tree (e.g., hands and upper limbs to InterMimic; hips, root, and stance legs to PHC), while maintaining global coherence and stability during contacts. 
In the experiments, our approach improves task success rates and contact stability on goal-conditioned dynamic HOI benchmarks (e.g., grasp-and-manipulate in dynamic motion styles) while maintaining motion realism and diversity on text-to-motion evaluations. Ablation studies show that both the prior blending in planning and composer-blended execution are essential, jointly accomplishing goal-driven diverse HOI tasks. Our contributions are: 
\begin{itemize}[leftmargin=*]
\item \textbf{Dynamic HOI Planning \& Execution.}
We proposed a novel two-stage framework that accomplishes goal-driven dynamic HOI:  
it generates kinematically dynamic, interaction-consistent motion plans and executes them robustly in a physics simulator.

\item \textbf{Prior Blending for Dynamic HOI Planning.}
A human motion diffusion model is guided by geometric/contact constraints from HOI priors, yielding planned motions that are simultaneously dynamic and interaction-consistent.

\item \textbf{Composer-Blended HOI Execution.}
A composer layer is introduced such that pretrained human and HOI imitation agents are spatio-temporally blended 
at action level, enabling robust execution of dynamic HOI behaviors that neither of the two agents succeeds alone. 
\end{itemize}
\section{Related Works}
\label{sec:Related}

\begin{figure*}[ht!]
  \centering
  \includegraphics[width=0.8\textwidth]{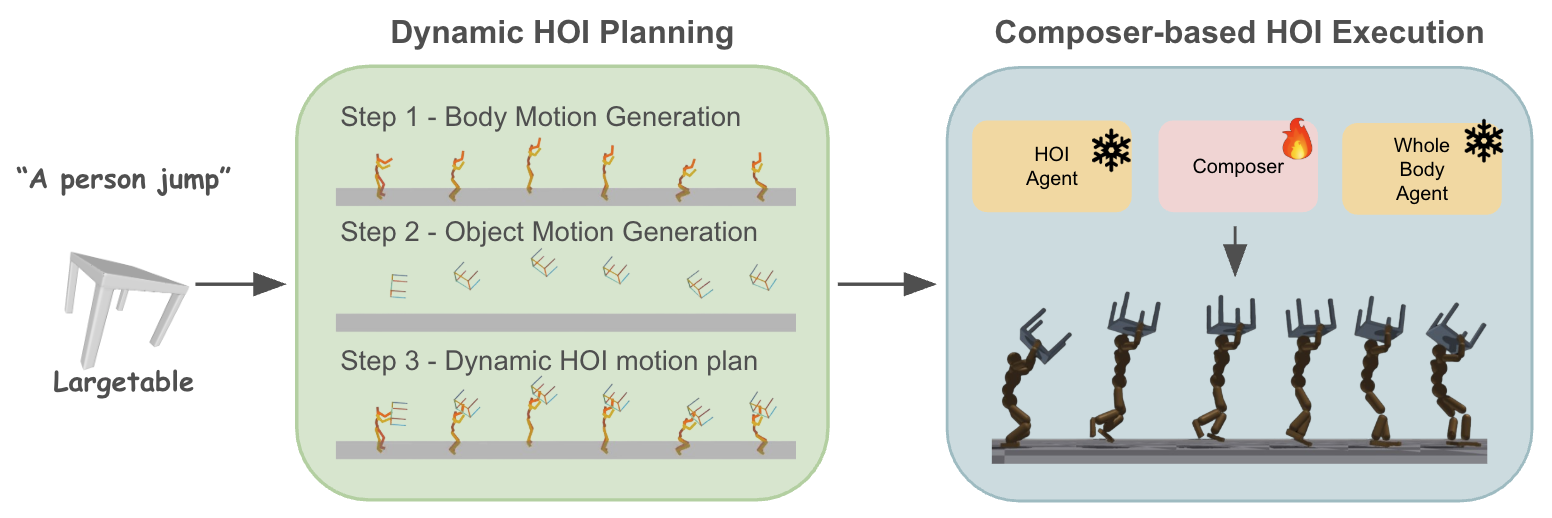}
  \caption{\textbf{Overall Framework.}
  The planning generates dynamic HOI sequences from a text prompt. The execution imitates these plans in a physics simulator by blending a versatile motion imitation agent and an HOI imitation agent utilizing their complementary skills.}
  \label{fig:overview}
\end{figure*}

\noindent \textbf{Diffusion-based Full-body and Human-Object Interaction Motion Generation.}
Large-scale human motion datasets \cite{Mahmood_2019_ICCV,Plappert2016,Guo_2022_CVPR,BABEL:CVPR:2021} and text encoders \cite{radford2021learning,devlin2019bert} have driven recent advances on text-to-motion generation. As in image generation~\cite{rombach2022high,ho2022imagen}, diffusion-based models have been successfully applied to text-to-motion generation \cite{tevet2023human,zhang2022motiondiffuse}. Various diffusion methods have been proposed for improved motion quality, trajectory-guided motion generation, and goal-conditioned
control~\cite{karunratanakul2024optimizing,chen2024taming,Zhao:DartControl:2025}. These methods rely purely on large-scale motion datasets without explicit physics constraints, often leading to penetration, floating, and foot sliding. PhysDiff~\cite{yuan2023physdiff} introduces physics-based constraints to mitigate such issues. They do not model object-aware conditioning and explicit contacts. Recent works \cite{wu2024thor,peng2025hoi,li2023controllable,xu2023interdiff,cong2025semgeomo,Geng_Hayder_Liu_Mian_2025,david,wu2025hoidynlearninginteractiondynamics} extend diffusion-based generation to HOI motions using HOI datasets \cite{bhatnagar22behave,GRAB:2020,zhao2024imhoi,lv2024himonewbenchmarkfullbody,zhang2023neuraldome,li2023object,xu2025interact} and conditioning object geometry, point clouds, or desired object trajectories. Especially, DAViD \cite{david} adapted a pretrained human motion diffusion model (MDM)~\cite{tevet2023human} via Low-Rank Adaptation (LoRA)~\cite{hu2022lora} to learn specific HOI concepts from few samples, while preserving the dynamics and diversity of MDM. However, most of the HOI diffusion models often yield sub-optimal and physically infeasible object contacts, due to the lack of temporal contact consistency and explicit contact supervision. Existing HOI datasets lack   dynamic human motions such as running, jumping, kicking, and dancing, while interacting with objects. 

\noindent \textbf{Humanoid Motion Imitation.}
Imitating human motions in physics environments has become a fundamental research area \cite{peng2018deepmimic,ren2023diffmimic,peng2019mcp,Luo2023PerpetualHC,traj2021}. Control policies are trained on a single reference motion \cite{peng2018deepmimic,ren2023diffmimic} or on a large-scale human motion dataset \cite{peng2019mcp,Luo2023PerpetualHC} using deep reinforcement learning (RL). Recent works such as PHC \cite{Luo2023PerpetualHC} can imitate versatile dynamic motions but struggle with HOI reference motions involving desired object trajectories.
On the other hand, efforts \cite{wang2024skillmimic,yu2025skillmimicv2,xu2025intermimic,graspnet2024} have been extended to HOI domains. They, however, are limited to domain-specific interactions \cite{wang2024skillmimic,yu2025skillmimicv2} or static interactions \cite{xu2025intermimic}, lacking robustness and scalability to diverse and dynamic HOI motions. 

\noindent \textbf{Vision-Language-Action~Models.}
The VLA framework integrates large-scale vision-language reasoning with embodied control systems. Models such as RT-2~\cite{pmlr-v229-zitkovich23a}, UniVLA~\cite{bu2025univla}, and GraspVLA~\cite{deng2025graspvla} employ transformer-based large language models to interpret textual instructions and visual observations and predict low-level robot actions. They demonstrate reasoning capabilities, yet limited to low degree of freedom (DoF)  manipulators. 
Helix~\cite{figure2025helix} extends VLA to humanoids but restricted to upper-body control, lacking full-body dynamics. HumanVLA~\cite{xu2024humanvla} trains a VLA student via teachers specialized to each object rearrangement tasks, but exhibits limited scalability to novel tasks, and remaining focused on static object rearrangement tasks. Humanoid-VLA~\cite{ding2025humanoid} employs a vision-language-conditioned MDM for motion planning, yet it focuses on walking, avoiding obstacles, and kicking a ball, which were simplified to goal-reaching tasks.

\noindent \textbf{Hierarchical Control Framework.}
High degree-of-freedom (DoF) humanoid control for multiple tasks remains challenging due to its dimensionality and instability and the instability in end-to-end learning. To address this, recent works typically address it through some paradigms including (1) skill embedding learning and (2) planning-execution frameworks. 
Several frameworks such as ASE~\cite{2022-TOG-ASE}, PULSE~\cite{luo2024universal}, and OmniGrasp~\cite{luo2024omnigrasp} learn skill embeddings from large-scale imitation learning. The skill embeddings, which capture transitions between adjacent timesteps, are then reused as an efficient exploration space of deep RL. They achieve robust locomotion and object grasping but lack scalability to new tasks. 
Other approaches \cite{ren2023insactor,tevet2025closd,lin2025simgenhoiphysicallyrealisticwholebody} proposed hierarchical planning-execution frameworks, where a high-level planner produces task-conditioned motion trajectories and a low-level controller executes the plan in physics environments. CLoSD~\cite{tevet2025closd} leveraged a diffusion planner \cite{chen2024taming} and PHC~\cite{Luo2023PerpetualHC} as a controller with task-specific finetuning, and achieved text-to-motion tasks such as locomotion, strike a target, sit down, and get up. 
However, in the target strike task, the desired trajectory of the object is not considered. 
Similarly, in the sit down and get up tasks, the sofa is fixed, meaning the framework is trained for static, scene-conditioned setups rather than dynamic HOI setups. They are not readily applicable to more challenging dynamic HOI tasks such as object carrying in desired  trajectories via consistent contacts. 

\noindent \textbf{Blending with pretrained Expert Models.}
To tackle complex tasks, recent works leverage pretrained expert models that have been trained on similar tasks, rather than learning from scratch. Residual Policy Learning (RPL)~\cite{silver2018residual,zhang2024residual,10650465,dexterous2020} and Progressive Neural Networks (PNN)~\cite{rusu2016progressive} enable adaptation of an existing model using additional neural networks. 
In contrast, Mixture of Experts (MoE)~\cite{shazeer2017outrageously,fedus2022switch,modular2022} allows dynamic selection or combination of multiple expert models for new tasks, effectively leveraging the capabilities of existing experts. While MoE \cite{shazeer2017outrageously} typically uses sparse gating or hard selection, recent approaches \cite{fedus2022switch} explore soft combinations of expert actions. Similarly, the composer network proposed in PHC~\cite{Luo2023PerpetualHC} dynamically determines the weights for each expert.

\section{Dynamic HOI Planning and Execution}
\label{sec:Method}



Our goal is to generate physically plausible and dynamic human–object interaction (HOI) motions. While previous works such as CLoSD~\cite{tevet2025closd} introduced a planning–execution framework for simple HOI tasks, it is not directly applicable to our problem. 
We introduce a novel planning-execution framework as illustrated in Fig.~\ref{fig:overview}. In Sec.~\ref{subsec:planning}, to synthesize dynamic HOI samples that preserve motion diversity while ensuring consistent hand-object contacts, we blend two complementary motion priors: (i) diverse and dynamic human motion prior from a pretrained MDM~\cite{tevet2023human} using AMASS~\cite{Mahmood_2019_ICCV} and (ii) HOI prior from FullBodyManip~\cite{li2023object}. In Sec.~\ref{subsec:execution}, we use the synthesized samples to train a composer-based imitation agent, which blends a HOI and whole-body controller in a physics simulator. 
Through this two-stage design, our approach simultaneously achieves physically plausible dynamic HOI motion generation and a robust imitation agent capable of executing such motions in physics environments.

\begin{figure}[t]
  \centering
\includegraphics[width=\columnwidth]{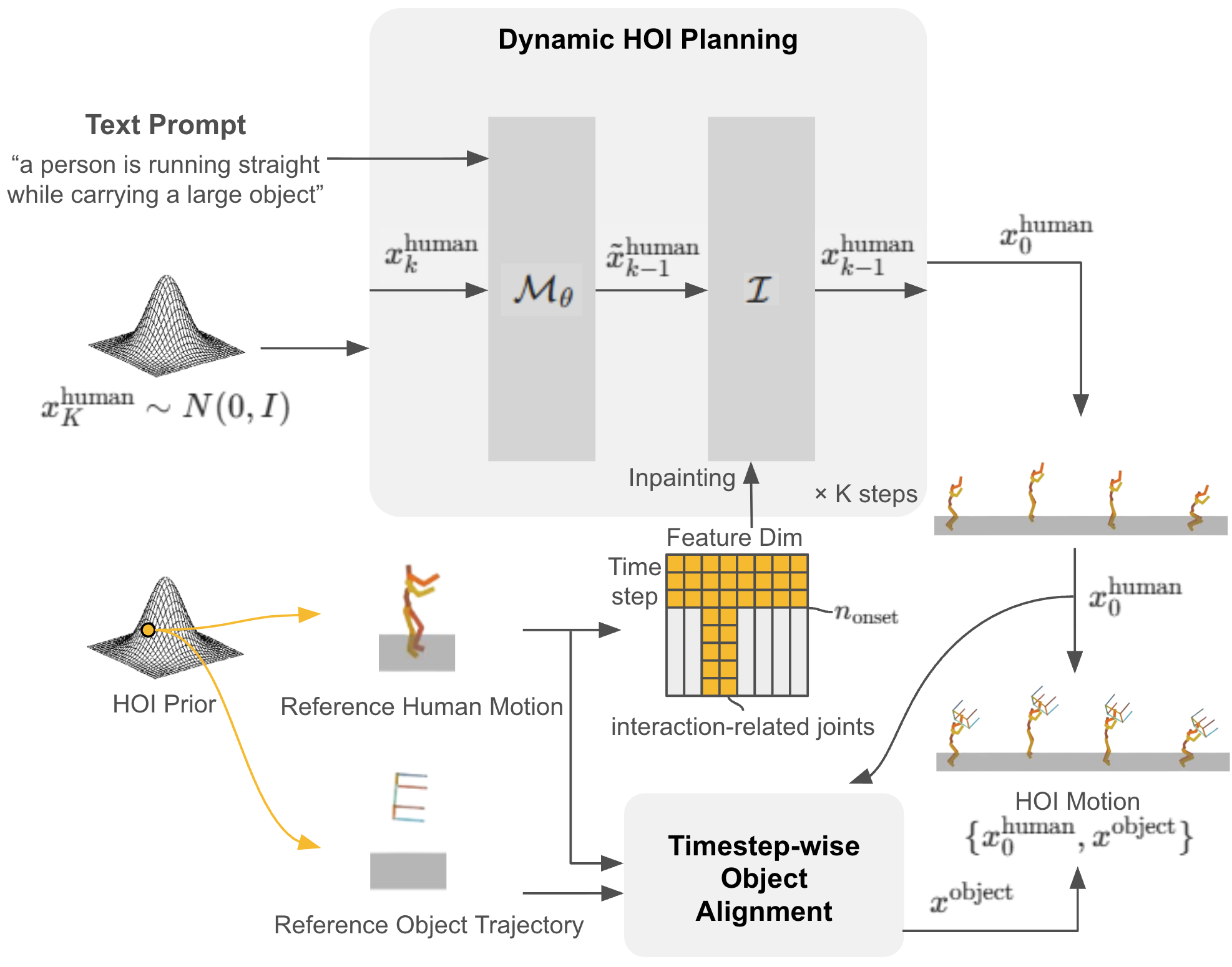}
  \caption{\textbf{Dynamic HOI Planning.}
  We sample human motions with MDM~\cite{tevet2023human} and inject HOI motion prior~\cite{li2023object} during diffusion sampling: full-body injection before the interaction onset $n_\text{onset}$ and interaction-related joints after $n_\text{onset}$.}
  \vspace{-3mm}
  \label{fig:stage1}
\end{figure}

\subsection{Prior Blending for Dynamic HOI Planning}
\label{subsec:planning}
We propose a scalable synthesis approach to obtain diverse and dynamic HOI motions $\mathbf{x}^{\text{human}} \in \mathbb{R}^{N \times D}$ and $\mathbf{x}^\text{object} \in \mathbb{R}^{N \times 7}$, without costly new data collection, where $N$ denotes motion length, $D$ denotes motion feature dimension, and $7$ represents the 3D position and quaternion of an object. 
As illustrated in Fig.~\ref{fig:stage1}, blending of MDM~\cite{tevet2023human} motion prior with FullBodyManip~\cite{li2023object} HOI motion prior is done by applying interaction-consistent guidance during MDM sampling. We detect the interaction onset, the first timestep that hand-object contact occurs, from FullBodyManip~\cite{li2023object}, and preserve its hand-object interaction consistently while generating the other body parts as follows: (i) before the onset, the full-body inpainting stabilizes initial human motions; (ii) after the onset, only interaction-related joints (e.g., thorax, shoulder, elbow, and wrist) are inpainted to maintain consistent hand-object contacts. This yields dynamic human motions with coherent hand-object interactions. Compared to LoRA~\cite{hu2022lora} adaptation, our method explicitly enforces contact consistency during the diffusion sampling. 

Before MDM denoising, we sample a single HOI motion clip from FullBodyManip, which denotes a sample from HOI motion prior. Using the contact mask, we detect the interaction onset step $n_\text{onset}$. We inpaint body poses $\mathbf{q}^\text{ref} \in \mathbb{R}^{|\mathcal{J}| \times 3}$ indicating the axis-angle rotation of body joints $\mathcal{J} = \{ \text{pelvis}, \text{left hip}, \ldots, \text{right wrist} \}$ defined by SMPL~\cite{SMPL:2015}. 
We propose an inpainting strategy that imputes $\mathbf{q}^\text{ref}$ into the denoised pose $\mathbf{q}$. For each joint $j$ at timestep $n$, inpainted pose $\hat{\mathbf{q}}_j[n]$ is calculated as
\begin{equation}
\mathbf{\hat{q}}_j[n] =
\begin{cases}
\mathbf{q}_j^{\text{ref}}[n] & n < n_{\text{onset}},\\
\mathbf{q}_j^{\text{ref}}[n_{\text{onset}}] & n \ge n_{\text{onset}} \text{ and } j \in \mathcal{J}_{\text{int}},\\
\mathbf{q}_j[n] & n \ge n_{\text{onset}} \text{ and } j \notin \mathcal{J}_{\text{int}},
\end{cases}
\end{equation}
where $\mathcal{J}_{\text{int}}$ denotes the interaction-related joints.
Let $\mathcal{M_\theta}$ denote a pretrained MDM parameterized with $\theta$, and $\boldsymbol{c}$ denotes text prompt embedding. Our denoising process at step $k$ with inpainting operator $\mathcal{I}$ is formulated as
\begin{equation}
\mathbf{\tilde{x}}_{k-1}^\text{human} = \mathcal{M}_\theta(\mathbf{x}_k^\text{human}, \mathbf{c}, k), \space \space \space
\mathbf{x}_{k-1}^\text{human} = \mathcal{I}(\mathbf{\tilde{x}}_{k-1}^\text{human}, \mathbf{q}^\text{ref}),
\end{equation}
where $\mathcal{I}$ overwrites the pose $\tilde{\mathbf{q}}_{k-1}^\text{human}$ of motion $\tilde{\mathbf{x}}_{k-1}^\text{human}$ with the inpainted pose $\hat{\mathbf{q}}_{k-1}^\text{human}$.
This inpainting strategy enforces interaction-consistent poses for interaction-related joints, while allowing the others to follow the diverse and dynamic motion prior of MDM. It yields stable pre-contact motions and coherent post-contact interactions, which is feasible for the composer training in Sec.~\ref{subsec:execution}.

After the human motion generation, we calculate the corresponding object trajectory via timestep-wise alignment. We first define an arbitrary set of contact anchors $\mathbf{A}^{\text{ref}}_{\text{(object)}} \in \mathbb{R}^{P \times 3}$ in the object local coordinate system, where $P$ denotes the number of anchor points. These anchors are transformed into hand local coordinate at onset step $n_{\text{onset}}$ to obtain $\mathbf{A}^{\text{ref}}_{\text{(hand)}} \in \mathbb{R}^{P \times 3}$, which represents desired anchor positions in the local coordinate system. For each step $n > n_\text{onset}$, the inpainted human motion provides the hand pose, through which we convert $\mathbf{A}^{\text{ref}}_{\text{(hand)}}$ back to the global coordinate to obtain the desired anchor positions $\mathbf{A}[n] \in \mathbb{R}^{P \times 3}$. The object's global-coordinate rotation $\textbf{R}^{\text{object}}[n] \in SO(3)$ and translation $\textbf{t}^{\text{object}}[n] \in \mathbb{R}^3$ are then estimated by solving Kabsch-SVD alignment:
\begin{equation}
(\textbf{R}^{\text{object}}[n], \textbf{t}^{\text{object}}[n]) = \arg\min_{\textbf{R},\textbf{t}} \big\| \textbf{RA}^{\text{ref}}_{\text{object}} + \textbf{t} - \textbf{A}[n] \big\|_F^2.
\end{equation}
Repeating this alignment for all steps after $n_\text{onset}$ yields $\mathbf{x}^\text{object} = \{ \textbf{R}^{\text{object}}[n], \textbf{t}^{\text{object}}[n]\}_{n=0}^{N-1}$. The resulting human-object motion pairs $\{ \textbf{x}_0^{\text{human}}, \textbf{x}^{\text{object}} \}$  serve as an imitation goals for the subsequent composer-based policy learning (Sec.~\ref{subsec:execution}).

\begin{figure}[t]
  \centering
\includegraphics[width=0.8\columnwidth]{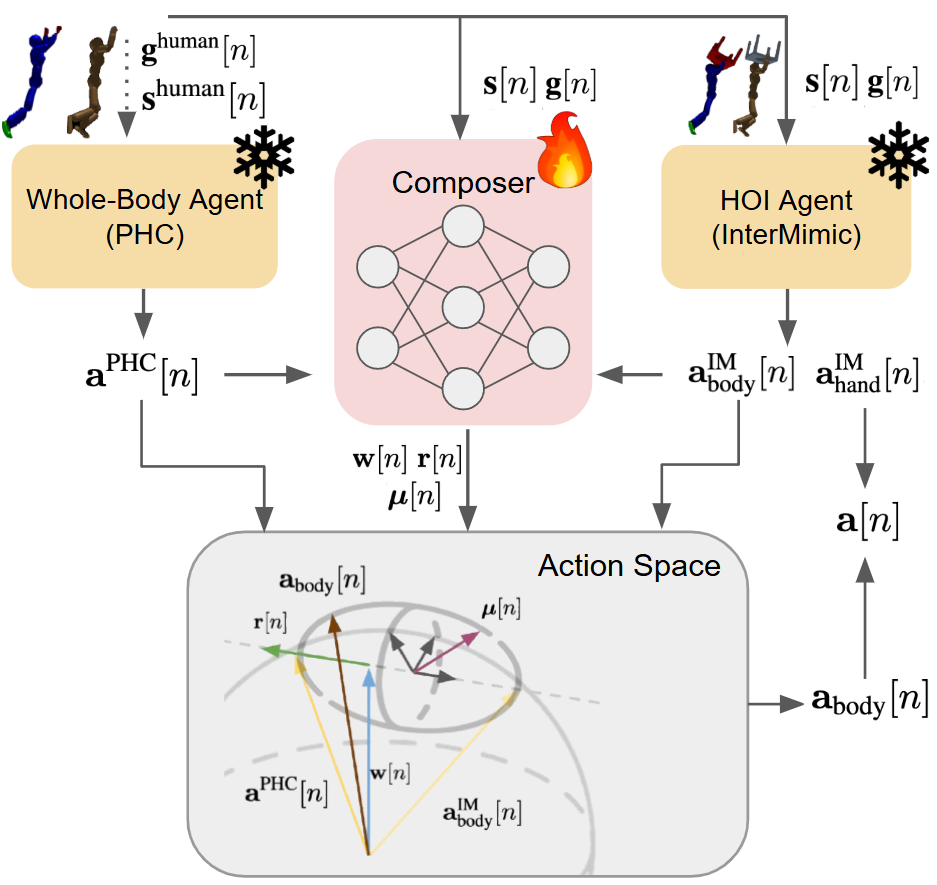}
  \caption{\textbf{Composer-based HOI Execution.}
  To track per-timestep goals from Sec.~\ref{subsec:planning}, we utilize two pretrained experts. The whole-body agent provides robust, dynamic full-body control, while the HOI agent provides contact-aware interaction behaviors including hands. A lightweight composer blends actions from two experts using per-DoF weights $\mathbf{w}[n]$ and $\mathbf{r}[n]$, 
  while encouraging exploration in additional subspace orthogonal to $\Delta \mathbf{a}_{\text{body}}[n]$ using $\mu[n]$.
  }
  \label{fig:stage2}
\end{figure}
\subsection{Composer-based HOI Execution}
\label{subsec:execution}

Our goal is to imitate dynamic HOI motions planned in Sec.~\ref{subsec:planning}, which demands an agent capable of both dynamic full-body motions and precise object interactions. To the best of our knowledge, none of the existing imitation experts can achieve it. To this end, we adopt an action-space blending framework that integrates two pretrained experts: (i) PHC~\cite{Luo2023PerpetualHC} for high-dynamics whole-body imitation, and (ii) InterMimic~\cite{xu2025intermimic} for contact-aware HOI imitation. Our unified imitation policy with the composer network blends their complementary capabilities in a spatio-temporal manner, enabling coherent and physically valid dynamic HOI imitation.
At each control step $n$, the physical environment provides observations $\textbf{s}[n] = \{ \textbf{s}^{\text{human}}[n], \textbf{s}^{\text{object}}[n]\}$, where $\textbf{s}^{\text{human}}[n]$ and $\textbf{s}^{\text{object}}[n]$ are the humanoid and object proprioception states, respectively. From the motion planned from Sec.~\ref{subsec:planning}, we define the reference in the same form as $\textbf{g}[n] = \{ \textbf{g}^{\text{human}}[n], \textbf{g}^{\text{object}}[n] \}$, where $\textbf{g}^{\text{human}}[n]$ and $\textbf{g}^{\text{object}}[n]$ denotes the reference humanoid and object states, respectively. These reference goals are used to compute imitation rewards for composer training.
Two frozen imitation experts are queried at every timestep. PHC receives the humanoid observations $(\textbf{s}^{\text{human}}[n], \textbf{g}^{\text{human}}[n])$, and outputs body-only actions $\textbf{a}^{\text{PHC}}[n] \in \mathbb{R}^{D_{\text{body}}}$, excluding hand joints. InterMimic takes the full observations $(\textbf{s}[n], \textbf{g}[n])$ and produces full-body actions $\textbf{a}^{\text{IM}}[n] \in \mathbb{R}^{D_{\text{full}}}$ including hand joints. ($D_{\text{full}}=D_{\text{body}}+D_{\text{hand}}$)

We blend body actions from the two experts and take hand actions directly from InterMimic, since PHC does not control hand joints. We use a lightweight eigenbasis composer that linearly combines the expert actions and adds a eigen vector base exploration in a low-dimensional subspace as illustrated in Fig.~\ref{fig:stage2}.
The composer predicts: (i) per-DoF interpolation weight $\mathbf{w}[n]\in[0,1]^{D_{\text{body}}}$, (ii) per-DoF bounded extrapolation weight $\mathbf{r}[n]\in[-\rho,\rho]^{D_{\text{body}}}$, and (iii) $S$ coefficients $\boldsymbol{\mu}[n]\in [-\sigma, \sigma]^S$ for the additional exploration basis $\mathbf{U}\in\mathbb{R}^{D_{\text{body}}\times S}$ ($\rho$ and $\sigma$ are hyperparameters).
To get the basis $\mathbf{U}$, we update a buffer that stores recent action differences between two experts, $\Delta \mathbf{a}_{\text{body}}[n] \;=\; \mathbf{a}^{\text{IM}}_{\text{body}}[n] - \mathbf{a}^{\text{PHC}}[n]$, then apply PCA on the buffer. The resulting top-$S$ basis $\textbf{U}$ denotes the eigenvectors of the exploration space, consists of bases orthogonal to each other. This strategy preserves a stable linear blending using $\mathbf{w}$ and $\mathbf{r}$, and adds low-dimensional orthogonal space exploration using $\mathbf{U}$ and $\boldsymbol{\mu}$ that improves efficient exploration while avoiding over-excitation in low-variance directions.
We define the final action as
\begin{equation}
\mathbf{a}_{\text{body}}[n] = \mathbf{a}^{\text{PHC}}[n] + \big(\mathbf{w}[n] + \mathbf{r}[n]\big)\odot \Delta \mathbf{a}_{\text{body}}[n] + \mathbf{U}\boldsymbol{\mu}[n],
\end{equation}
\begin{equation}
\mathbf{a}_{\text{hand}}[n] = \mathbf{a}^{\text{IM}}_{\text{hand}}[n],
\end{equation}
\begin{equation}
\mathbf{a}[n] = \big[\mathbf{a}_{\text{body}}[n];\,\mathbf{a}_{\text{hand}}[n]\big] \in \mathbb{R}^{D_{\text{full}}}.
\end{equation}
In the training time, the composer learns to predict $\mathbf{w}[n]$, $\mathbf{r}[n]$, and $\boldsymbol{\mu}[n]$ that best imitate the reference motion planned in Sec.~\ref{subsec:planning}. Training the composer network adopts the standard imitation objective of InterMimic~\cite{xu2025intermimic}. The reward penalizes deviations of joint and object states from the planned reference motion, ensuring that hand–object interactions remain consistent in the simulator. PHC and InterMimic remain frozen during the composer training, so their respective strengths are preserved. By learning per-timestep, per-joint weights, the composer enables spatio-temporally compositional HOI execution that neither of the experts alone consistently succeeds.

\begin{table*}[t]
  \centering
  \caption{
  \textbf{Quantitative Comparisons on Dynamic HOI Generation.} 
  `HOI' shows whether the method generates both human and object motions. `Physics' indicates whether the method exploits physics-based simulation for motion generation.
  }
  \label{tab:stage1-result}
  \vspace{2pt}
  \setlength{\tabcolsep}{5pt}
  \scalebox{0.85}{
  \begin{tabular}{lcccccccccc}
    \toprule
    \multirow{2}{*}{Method} & \multirow{2}{*}{HOI} & \multirow{2}{*}{Physics} & \multicolumn{2}{c}{\textsc{HOI quality}} & \multicolumn{4}{c}{\textsc{Physical plausibility}} & \multicolumn{2}{c}{\textsc{Motion quality}} \\
    \cmidrule(lr){4-5}\cmidrule(lr){6-9}\cmidrule(lr){10-11}
        & & & $C_{\%}$ $\uparrow$ & $C_\text{cons}$ $\downarrow$
        & $\text{Pene}_\text{obj}$ $\downarrow$ & Skate $\downarrow$ & Float $\downarrow$ & $\text{Jitter}_\text{pos}$ $\downarrow$ 
        & R-Prec $\uparrow$ & Diversity $\uparrow$\\
    \midrule
    MDM         &  &  & -- & -- & -- & \textbf{0.133} & 29.3 & $9.43 \times 10^4$ & \textbf{0.374} & \textbf{7.61} \\
    HOI-Diff    & \checkmark &  & 0.285 & 18.2 & 8.18 & 0.633 & \underline{17.8} & $\boldsymbol{1.56 \times 10^4}$ & 0.257 & 4.66 \\
    DAViD       & \checkmark &  & 0.848 & 10.9 & 4.842 & 0.261 & 20.4 & $6.69 \times 10^4$ &  0.310 & \underline{6.70}  \\
    \midrule
    $\text{Ours}_\text{P}$    & \checkmark &  & \textbf{1.000} & \textbf{0.906} & \underline{4.196} & \underline{0.217} & 30.1 & $5.93 \times 10^4$ & \underline{0.332} & 5.56  \\
    $\text{Ours}_\text{P+E}$  & \checkmark & \checkmark  & \underline{0.999} & \underline{2.95} & \textbf{0.009} & 0.786 & \textbf{9.95} & $\underline{4.53 \times 10^4}$ & 0.316 & 3.97 \\
    \bottomrule
  \end{tabular}
  }
  \vspace{2pt}
\end{table*}

\section{Experiments}
\label{sec:Experiments}

\noindent \textbf{Humanoid Kinematics.}
We use the same humanoid agent as in InterMimic~\cite{xu2025intermimic}, consisting of 51 joints defined by SMPL-X~\cite{SMPL-X:2019}—21 body joints excluding the root and 15 joints for each hand—each having 3 degrees of freedom, resulting in a 153-DoF action space. Each joint is actuated by a PD controller to ensure stable and smooth control. All training and evaluation are conducted in the Isaac Gym~\cite{makoviychuk2021isaac}, following PHC~\cite{Luo2023PerpetualHC} and InterMimic~\cite{xu2025intermimic}.

\noindent \textbf{Dynamic Motion Style.}
We evaluate our method for four dynamic motion styles—(i) \emph{run forward}, (ii) \emph{jump forward}, (iii) \emph{high kick}, and (iv) \emph{dance}—which are absent from FullBodyManip~\cite{li2023object}, and at the same time, are feasible under hand–object contact. This choice probes out-of-distribution dynamics while ensuring that each style admits meaningful HOI interactions.

\noindent \textbf{Interaction Category.}
To assess robustness to object properties and contact modality, we evaluate our method for three objects—\emph{smallbox} and \emph{largetable} which involve two-handed interactions with different sizes, and \emph{clothesstand} which involves one-handed interaction. Using these objects, we define five interaction categories—(i) \emph{carry a largetable}, (ii) \emph{lift a largetable overhead}, (iii) \emph{carry a smallbox}, (iv) \emph{carry a clothesstand with the left hand}, and (v) \emph{carry a clothesstand with the right hand}. This setup demonstrates the robustness of our agent over object sizes and shapes, and one-handed and two-handed contacts.



\begin{figure}[t]
  \centering
  \begin{subfigure}[t]{0.49\columnwidth}
    \centering
    \includegraphics[width=\columnwidth]{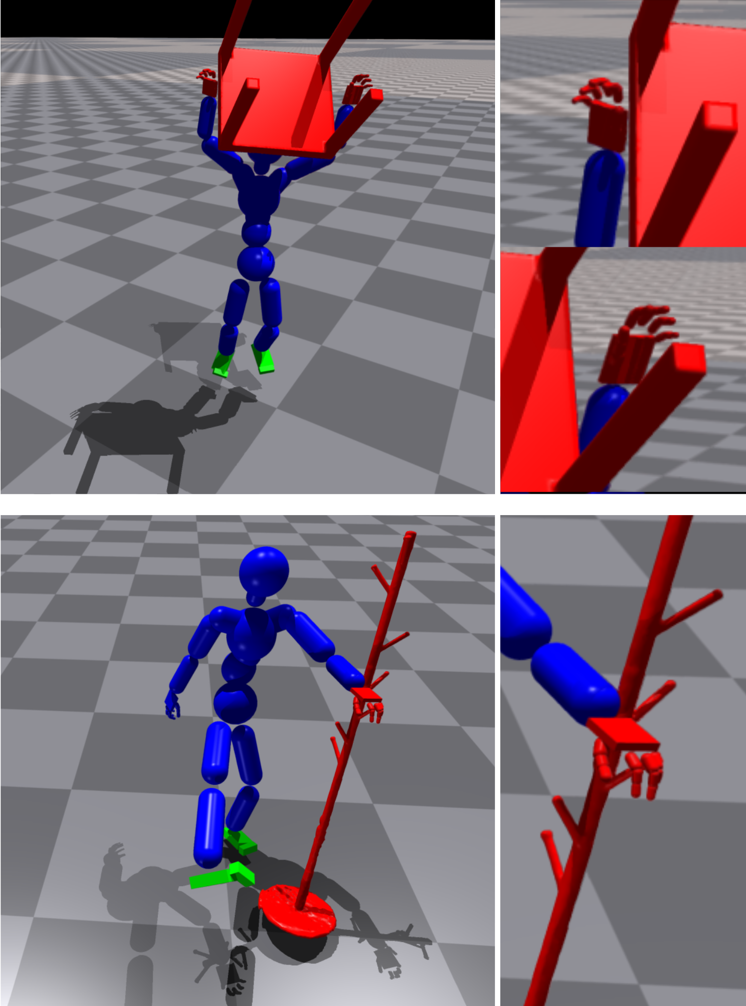}
    \caption{DAViD~\cite{david}}
  \end{subfigure}\hfill
  \begin{subfigure}[t]{0.49\columnwidth}
    \centering
    \includegraphics[width=\columnwidth]{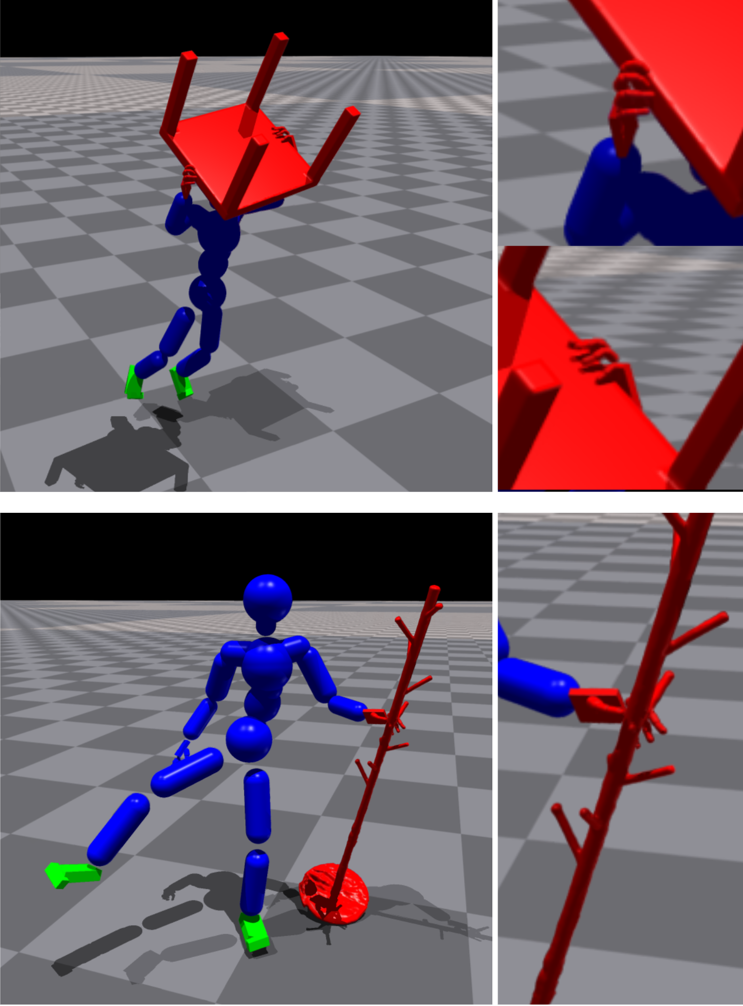}
    \caption{$\text{Ours}_\text{P}$}
  \end{subfigure}
  \caption{\textbf{Comparisons on Dynamic HOI Planning.}
  Contacting hands and objects are shown in red. 
  $\text{Ours}_\text{P}$ produces more accurate and temporally consistent hand-object interactions compared to DAViD~\cite{david}, benefiting from our step-wise alignment even without the physics-based execution stage. }
  \label{fig:stage1_qualitative}
\end{figure}

\subsection{Text-to-Dynamic HOI Motion Generation}
\label{subsec:text-to-dynhoi}

\noindent \textbf{Datasets.}
Since no existing dataset provides dynamic human-object interactions, we construct a dedicated dynamic HOI test set by defining text descriptions that combine various motion styles (e.g., \textit{``a person runs straight"}, \textit{``a person jumps forward"}) with object interaction contexts (e.g., \textit{``while carrying a large object"}, \textit{``while carrying a long object with his left hand"}). Here, we use \textit{``objects"} instead of using their specific categories, since these are out-of-distribution of the text encoder by MDM.

\noindent \textbf{Metrics.}
We evaluate text-to-dynamic HOI generation in terms of HOI quality~\cite{li2023object}, physical plausibility~\cite{yuan2023physdiff}, and motion quality~\cite{tevet2023human}. \textbf{HOI quality} reports contact percentage~\cite{li2023object} ($C_\%$ $\uparrow$) that measures proportion of timesteps maintaining hand-object contact, and \emph{contact consistency} ($C_\text{cons}$ $\downarrow$), a metric we propose, which measures the temporal consistency of hand-object contact. $C_\text{cons}$ computes the standard deviation of the relative pose of the object vertices expressed in the local hand coordinate, where the lower value indicates a more consistent contact and inherently reflects both transitional and rotational consistency. \textbf{Physical plausibility}, following {PhysDiff}\cite{yuan2023physdiff}, measures \emph{foot skate} ($\downarrow$), body \emph{float} ($\downarrow$), and \emph{$\text{Jitter}_\text{pos}$} ($\downarrow$) of body positions. We also report object-floor penetration ($\text{Pene}_\text{obj}$ $\downarrow$) which measures the interpenetration depth between the object and the ground.
\textbf{Human motion quality} is assessed following \cite{tevet2023human} by Top-3 R-Precision ($\uparrow$) and Diversity ($\uparrow$) to measure text–motion alignments and variations. Frechet Inception
Distance (FID) is not measured due to the absence of GT dynamic HOI motions. 

\noindent \textbf{Baselines.}
We compare our dynamic HOI generation with: \textbf{MDM}~\cite{tevet2023human}, a text-to-motion model trained on AMASS~\cite{Mahmood_2019_ICCV} without explicit object interaction, \textbf{HOI-Diff}~\cite{peng2025hoi}, a text-to-HOI diffusion model, and \textbf{DAViD}~\cite{david}, which augments an pretrained MDM with interaction category-specific LoRA~\cite{hu2022lora} finetuning on FullBodyManip~\cite{li2023object} for dynamic HOI. \textbf{$\text{Ours}_\text{P}$} only utilizes our dynamic HOI planning stage (Sec.~\ref{subsec:planning}) and \textbf{$\text{Ours}_\text{P+E}$} fully utilizes our planning and execution stages. 
For a fair comparison, we use the same text prompts for all methods, corresponding to the dynamic motion styles and interaction contexts described above. For DAViD, we append an object-specific suffix to each prompt, as its LoRA was finetuned with the convention. In addition, as in DAViD, evaluations are performed only on timesteps where the human is in contact with the object. HOI-quality metrics are excluded for MDM since it does not generate object trajectories.

\noindent \textbf{Results.} 
In terms of HOI quality, both our methods $\text{Ours}_\text{P}$ and $\text{Ours}_\text{P+E}$ achieve the best $C_\%$ and $C_\text{cons}$, showing more stable and consistent hand-object interactions compared to other baselines. As shown in Fig.~\ref{fig:stage1_qualitative}, DAViD often fails to maintain stable hand-object alignment, leading to visible gaps and inconsistent grasping poses, while our inpainted motion generation and step-wise object alignment preserve accurate and temporally coherent hand-object relative positions and orientations throughout the interaction. 
$\text{Ours}_\text{P+E}$ exhibits slightly higher $C_\text{cons}$ than $\text{Ours}_\text{P}$, which is an expected result of physically simulated interactions where minor slips and rotations naturally occur. 
For physical plausibility, $\text{Ours}_\text{P+E}$ effectively reduces artifacts such as $\text{Pene}_\text{obj}$, float, and $\text{Jitter}_\text{pos}$. While $\text{Ours}_\text{P}$ records the lowest skate score, it represents an idealized motion without physical validation. In contrast, $\text{Ours}_\text{P+E}$ shows a higher skate score due to physically valid corrective foot motions for balance recovery during dynamic HOI, which are natural rather than artifacts. 
In human motion quality evaluation, since our experiments involve a smaller number of motions and dynamic HOI-specific text descriptions that differ from those in AMASS~\cite{Mahmood_2019_ICCV}, we focus on relative comparisons across HOI methods. 
The inpainting strategy using consistent HOI motion prior reduces the diversity by constraining interaction-related joints. However, this trade-off between the diversity and accurate hand-object contact is supported by our higher R-Precision compared to DAViD, suggesting better alignment between dynamic HOI text descriptions and the generated motions.

\begin{table}[t]
  \centering
  \caption{
  \textbf{Quantitative Comparisons on Dynamic HOI Imitation.}
  $SR$ denotes the average success rate across five motion styles, evaluated with their respective style-specific goals.}
  \label{tab:stage2-result}
  \vspace{2pt}
  \setlength{\tabcolsep}{4.5pt}
  \scalebox{0.85}{
  \begin{tabular}{lcc|ccc}
    \toprule
    Method & $SR$ $\uparrow$ & T & $D$ $\uparrow$ & $E_{\text{HOI}}$ $\downarrow$ & $\text{Jitter}_\text{DoF}$ $\downarrow$\\
    \midrule
    PPO                                 & 0.227 & 75 & 0.859 & 59.098 & $0.424 \times 10^3$\\
    \midrule
    PHC~\cite{Luo2023PerpetualHC}       & 0.397 & - & 1.166 & 12.965 & $1.350\times 10^3$\\
    $\text{PHC}_\text{R}$                     & 0.313 & 15 & 1.834 & 26.024 & $1.762\times 10^3$\\
    \midrule
    InterMimic~\cite{xu2025intermimic}  & 0.376 & - & 2.407 & 21.379 & $0.436\times 10^3$\\
    $\text{InterMimic}_\text{R}$              & 0.310 & 15 & 2.324 & 34.675 & $\boldsymbol{0.336\times 10^3}$\\
    $\text{InterMimic}_\text{FT}$            & \underline{0.526} & 75 & \textbf{7.37} & \textbf{8.635} & $0.370\times 10^3$\\
    \midrule
    Ours                                & \textbf{0.591} & 23 & \underline{4.607} & \underline{11.667} & $\underline{0.359\times 10^3}$\\
    \bottomrule
  \end{tabular}
  }
\end{table}

\begin{figure}[t]
  \centering
  \makebox[\columnwidth][c]{%
    \begin{subfigure}[t]{0.47\columnwidth}
      \centering
      \includegraphics[width=\linewidth]{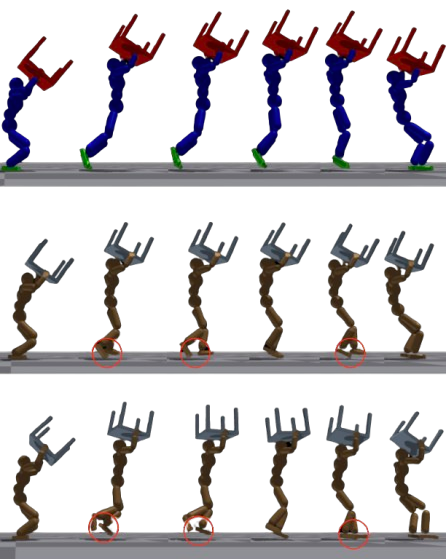}
      \caption{Frame-wise snapshots}
      \label{fig:qual_frame}
    \end{subfigure}
    \hspace{0.02\columnwidth}
    \begin{subfigure}[t]{0.47\columnwidth}
      \centering
      \includegraphics[width=\linewidth]{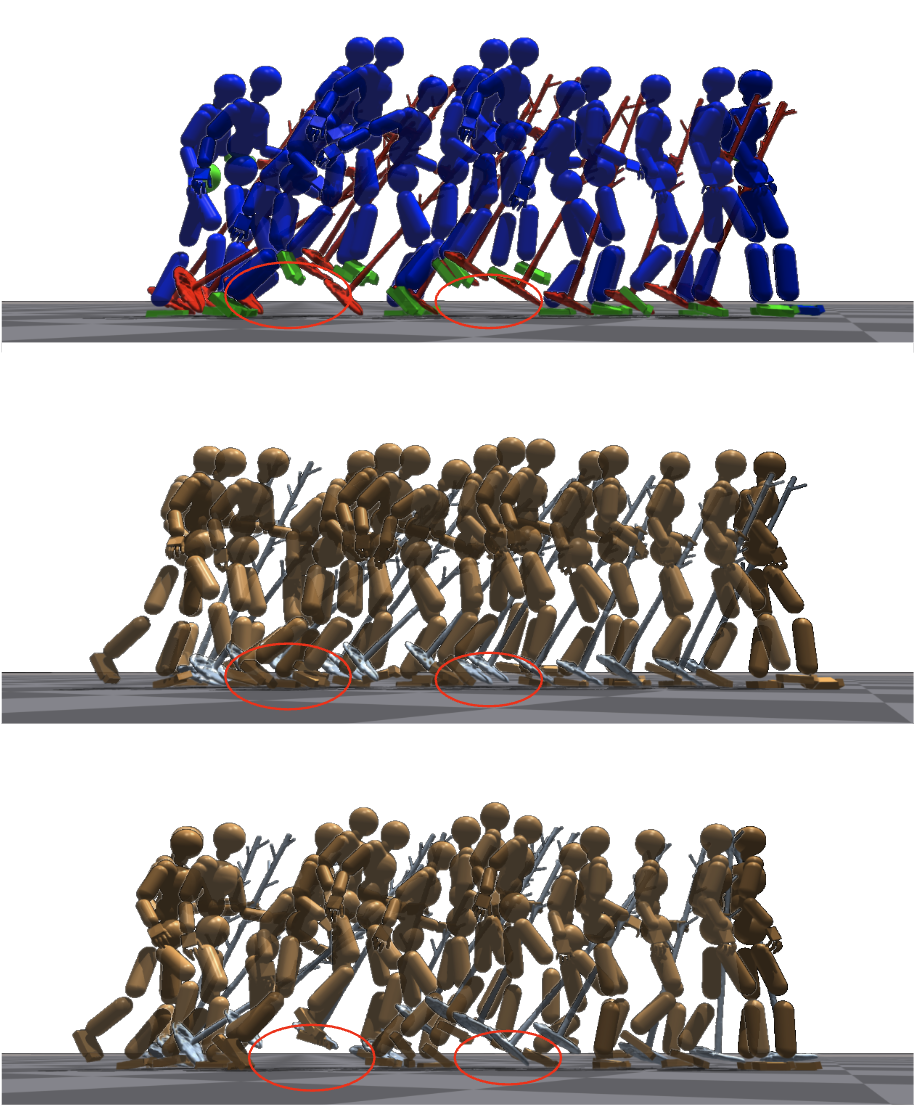}
      \caption{Trajectory overlays}
      \label{fig:qual_traj}
    \end{subfigure}
  }
  \caption{\textbf{
  Comparisons on dynamic HOI Imitation.}
  Reference motion (top row), motion from $\text{InterMimic}_\text{FT}$ (middle row), and motion from Ours (bottom row) for a text prompt \textit{``A person jumps, a largetable"}.
  The two visualizations, (a) and (b), compare per-frame alignment and the overall trajectory for distinct references. 
  As shown, our method successfully executes the dynamic jump to follow the reference, whereas $\text{InterMimic}_\text{FT}$ often collapses to a local optimum—short shuffling steps that minimize falling—rather than preserving the dynamic motion style.}
  \label{fig:stage2_qualitative}
\end{figure}

\subsection{Dynamic HOI Imitation in Physics Simulator}
\label{subsec:imitation}

\noindent \textbf{Dynamic HOI Tasks.}
\label{par:dynamic-hoi-tasks}
We consider tasks combining diverse motion styles and interaction categories, to measure robustness. An episode is regarded as success if (i) the motion style-specific goal is achieved and (ii) human-object interaction is maintained without a drop while following the HOI motion. Detailed task setups and motion style-specific goals are described in the supplementary. 

\noindent \textbf{Metrics.}
We report \emph{Success Rate} ($SR\uparrow$) which indicates the fraction of episodes that meet the style-conditioned success criterion, and \emph{Training Time} (T) until the convergence. We also report HOI imitation metrics such as \emph{Mean Duration} ($D\uparrow$), and \emph{HOI Error} ($E_{\text{HOI}}\downarrow$) over positions of 22 SMPL joints and the object following InterMimic~\cite{xu2025intermimic}. We further report \emph{$\text{Jitter}_\text{DoF}$} to measure smoothness. $E_\text{HOI}$ is measured after the alignment of the roots in $n=0$.

\noindent \textbf{Baselines.}
We select baselines to reflect the two most common ways to adapt pretrained controllers to new tasks—\emph{finetuning} and \emph{residual adaptation}—and to relate directly to our action-space blending. Since PHC~\cite{Luo2023PerpetualHC} does not control hands, we only evaluate its residual adaption to include hand actions.
Using dynamic HOI motions planned in Sec.~\ref{subsec:planning} as a training set, we compare our method with \textbf{PPO}~\cite{schulman2017proximal} trained from scratch, pretrained \textbf{PHC}, PHC with residual adaptation (\textbf{$\text{PHC}_\text{R}$}), pretrained \textbf{InterMimic}~\cite{xu2025intermimic}, and InterMimic with residual adaptation (\textbf{$\text{InterMimic}_\text{R}$}) and finetuning (\textbf{$\text{InterMimic}_\text{FT}$}). 

\noindent \textbf{Results.}
Tab.~\ref{tab:stage2-result} evaluates over our central hypothesis: a composer that blends pretrained experts enables dynamic, contact-rich HOI. We emphasize the evaluation on $SR$, since the goal is not only to avoid falls or drops but to execute the prompt-specific dynamic motions while maintaining contacts. Our method attains the highest $SR$, indicating the most reliable dynamic HOI among all methods.
The weak performance of PPO from scratch confirms that dynamic HOI is challenging to learn without priors, motivating a pretrained-expert approach. Simply attaching a residual head underperformed; in $\text{InterMimic}_\text{R}$, residuals add noise to hand actions after grasp formation, increasing drops, while in $\text{PHC}_\text{R}$, a non-hand controller cannot be patched into robust contact maintenance. Thus, the residuals remain ineffective for cross-domain skill transfer, rather a structured composition is required.

As a fidelity measure, we report $E_{\mathrm{HOI}}$, which directly reflects imitation accuracy. $\text{InterMimic}_\text{FT}$ yields lower geometric errors than ours, but qualitative results shown in Fig.~\ref{fig:stage2_qualitative} reveal the local optimum: to stay upright, the $\text{InterMimic}_\text{FT}$ replaces true lower-body dynamic behaviors, e.g. jumping in the air, with rapid shuffling steps. Since references are synthetic and do not account for object mass or induced lean/inertia, aggressively minimizing geometric error lead to less dynamic, less physically faithful motions. By contrast, our composer preserves contacts and produces the intended dynamic style, accepting a modest geometric deviation that better aligns with the task goal. The other quantitative metrics, $D$ and $\text{Jitter}_\text{DoF}$, also support our explanation. 
When dynamics are required, the $\text{InterMimic}_\text{FT}$ often substitutes rapid shuffling, increasing stance time and reducing falls inflating $D$ yet introducing frequent micro-corrections that raise $\text{Jitter}_\text{DoF}$. Our composer executes the intended dynamic phases while maintaining contact, achieving the $D$ and $\text{Jitter}_\text{DoF}$ comparable to baselines without sacrificing the target styles. Finally, our method is training efficient, reaches a superior $SR$ with roughly $3\times$ less time than $\text{InterMimic}_\text{FT}$. Overall, the results support that blending pretrained experts is a reliable and scalable route to dynamic, contact-rich HOI.



\subsection{Ablation Studies}
\label{subsec:ablation}
\begin{figure}[t]
  \centering
\includegraphics[width=0.45\columnwidth, height=0.4\columnwidth]{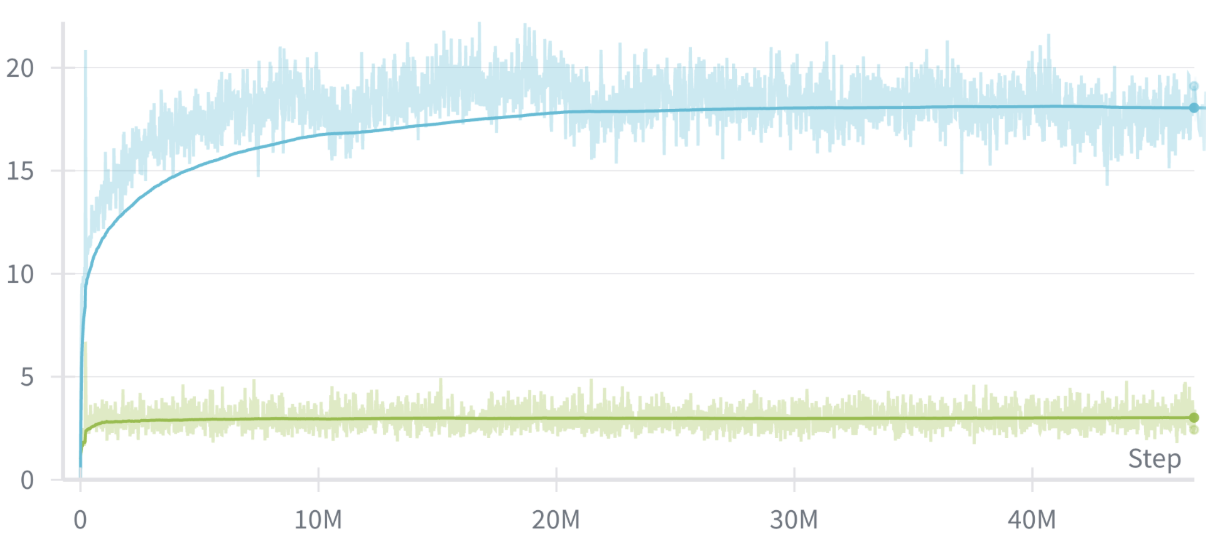}
\includegraphics[width=0.45\columnwidth, height=0.4\columnwidth]{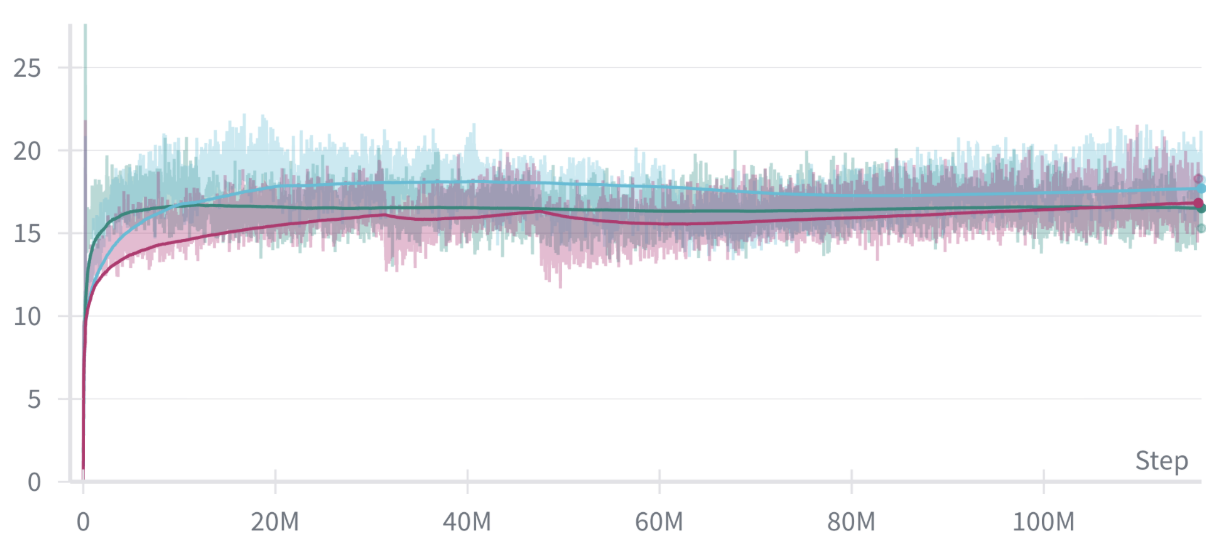}
  \caption{\textbf{Ablations.} \textbf{Left:} Learning curves of the composer trained on dynamic HOI motions generated by ours (blue) and DAViD (green). \textbf{Right:} Learning curves of the composer with different blending methods—$\text{Ours}_\text{MLP+PCA}$ (blue), $\text{Ours}_\text{MLP}$ (pink), and Hard MoE (dark green).}
  \label{fig:learning_curve}
\end{figure}

\noindent \textbf{Ablation in Planning.}
We study how the dynamic HOI planning stage affects the subsequent execution stage. We compare our prior blending-based dynamic HOI planning (Sec.~\ref{subsec:planning}) against DAViD~\cite{david}. Each variant is evaluated by how well our composer learns from the dynamic HOI motions generated by each planning strategy. 
As shown in Fig.~\ref{fig:learning_curve} (left), the composer trained with our planning  exhibits faster convergence and higher rewards, while DAViD suffers from unstable hand-object contacts, causing unstable composer training.

\noindent \textbf{Ablation in Execution.}
We analyze which blending strategy best adapts to dynamic HOI. 
We compare $\textbf{Heuristic}_\textbf{Hand}$ (InterMimic for hands, PHC for body), $\textbf{Heuristic}_\textbf{Arm}$ (InterMimic for hands+arms, PHC for the others), \textbf{Hard MoE} (single expert per timestep), \textbf{Hard MoE (Joint)} (extended to spatio-temporal switching),
\textbf{Ours\textsubscript{MLP}}, which learns per-DoF, per-step interpolation weights in action space 
but no subspace exploration, and \textbf{Ours\textsubscript{MLP+PCA}} described in Sec.~\ref{subsec:execution}.
Tab.~\ref{tab:ablation_execution} shows the Heuristic policies show the lowest $SR$, indicating that a fixed partition of responsibility is not enough; the controller must blend pretrained experts \emph{spatio–temporally}. Hard MoE improves the duration by allowing temporal switching, yet still shows low $SR$. 
Ours\textsubscript{MLP} already yields strong performance, while Ours\textsubscript{MLP+PCA} surpasses the others, which indicates PCA/eigenbasis-spanned exploration space increases the performances. A detailed comparison of the training curves is shown in Fig.~\ref{fig:learning_curve} (right).

\begin{table}[t]
\centering
\caption{
  \textbf{Ablation on Execution.}
  Each blending strategies except Heuristic were trained by dynamic HOI motions planned in Sec.~\ref{subsec:planning}.
}
\label{tab:ablation_execution}
\scalebox{0.85}{
\begin{tabular}{lccccc}
\toprule
Blending & SR$\uparrow$ & D$\uparrow$ & $E_\text{HOI}$$\downarrow$ \\
\midrule
Heuristic$_{\text{Hand}}$ & 0.365 & 1.069  & 13.04 \\
Heuristic$_{\text{Arm}}$ & 0.472 & 1.603  & 17.18 \\
\midrule
Hard MoE                 & 0.383 & 2.748 & 20.48 \\
Hard MoE (Joint)         & 0.383 & 2.535  & 26.45 \\
\midrule
$\text{Ours}_\text{MLP}$ & 0.571 & 3.034 & 18.03 \\
$\text{Ours}_\text{MLP+PCA}$ & \textbf{0.591} & \textbf{4.607} & \textbf{11.67} \\
\bottomrule
\end{tabular}
}
\end{table}

\section{Conclusions}
\label{sec:conclusion}
In this work, we introduced a novel framework that blends pretrained modular controllers from distinct motion domains to execute behaviors that neither controller can reliably perform alone. We first proposed a planning strategy that injects an interaction-consistent HOI prior into a human motion diffusion model, producing dynamic, contact-consistent HOI plans. Using these plans, we then trained a composer that spatio-temporally blends a dynamic full-body imitation agent with a static-HOI agent, yielding higher success rates with a substantially lower training cost. Beyond simple linear mixing, our composer explores a compact PCA/eigenbasis spanned subspace, enabling exploration while leveraging both priors’ action spaces effectively. Taken together, our results demonstrate an efficient path toward universal humanoid HOI control by reusing strong priors and learning only what are necessary to compose them.
For future work, we aim to orchestrate more pretrained agents and foundation models, and investigate physics-aware planning that accounts for motion dynamics and object properties including mass and contact geometry.  

\FloatBarrier

\noindent \textbf{Acknowledgments.}
This work was supported by 
NST grant (CRC 21015, MSIT), 
IITP grant (RS-2023-00228996, RS-2024-00459749, RS-2025-25443318, RS-2025-25441313, MSIT),
and KOCCA grant (RS-2024-00442308, MCST).
This work was also supported by MSIT (Ministry of Science and ICT), Korea, under the Graduate School of Metaverse Convergence support program (IITP-2022(2025)-RS-2022-00156435) supervised by the IITP (Institute for Information \& Communications Technology Planning \& Evaluation).

{
    \small
    \bibliographystyle{ieeenat_fullname}
    \bibliography{main}
}

\clearpage
\setcounter{page}{1}
\setcounter{section}{0}
\setcounter{table}{0}
\setcounter{figure}{0}
\renewcommand\theHsection{supp.\Alph{section}}
\renewcommand\theHsubsection{supp.\Alph{section}.\arabic{subsection}}
\renewcommand\thesection{\Alph{section}}
\renewcommand\thefigure{\Alph{figure}}
\renewcommand\thetable{\Alph{table}}
\maketitlesupplementary

\section{Supplementary Video}
\label{sec:supp_demo_video}

The submitted video qualitatively illustrates how our framework generates dynamic and physically valid HOI motions, comparing its planning and execution with multiple baselines. 

\begin{itemize}
\item \textbf{Prior-Blending for HOI Planning.} Under identical text prompts, $\textrm{Ours}_\textrm{P}$ generates HOI motion plans with more feasible hand–object contacts than baselines.

\item \textbf{Improving Physical Plausibility.} Physical artifacts in $\textrm{Ours}_\textrm{P}$ plans—such as object–ground penetration—are corrected in the composer-based execution stage.

\item \textbf{Composer-based Execution.} Our composer achieves the highest task success rate and the most stable HOI imitation among all baselines.

\item \textbf{More Qualitative Results.} Our full pipeline demonstrates dynamic and scalable HOI behaviors across a wide range of motion styles and object categories.
\end{itemize}

\section{Setup for Physics Simulation}
\label{sec:supp_phys_sim}

Our HOI simulation environment is built on Isaac Gym~\cite{makoviychuk2021isaac}, following InterMimic~\cite{xu2025intermimic} to convert the SMPL~\cite{SMPL:2015}-based human model from reference motions into rigid-body representations suitable for physics simulation. Each body part is approximated with box or cylinder primitives, while object meshes are converted using convex decomposition as illustrated in Fig.~\ref{fig:supp_objects}.

Although our framework utilizes both PHC~\cite{Luo2023PerpetualHC} and InterMimic~\cite{xu2025intermimic} as imitation experts, the two policies were originally trained under different physics parameter settings. To ensure consistency across all HOI imitation experiments, we adopt the physics parameters used in PHC as our default configuration. We verified that InterMimic maintains comparable performance under PHC's physics configuration. Consequently, all imitation experiments are conducted using the PHC physics parameters. Detailed physical configurations are provided in Tab.~\ref{tab:supp_sim_param}.

The joint Range of Motion (RoM) follows the biomechanical constraints specified by InterMimic~\cite{xu2025intermimic}. Finger flexion and extension are fully allowed to support natural grasping behaviors, while the Metacarpophalangeal joints are restricted to prevent finger interpenetration. Although PHC~\cite{Luo2023PerpetualHC} also limits the RoM of several body joints to avoid body interpenetration, we relax these constraints to preserve the capability of the InterMimic agent. The full RoM specifications for each joint are listed in Tab.~\ref{tab:supp_rom}.

\begin{table}[t]
  \centering
  \caption{
  Simulation parameters for Isaac Gym~\cite{makoviychuk2021isaac} used in this paper.}
  \label{tab:supp_sim_param}
  \scalebox{0.85}{
  \begin{tabular}{l|c}
    \toprule
    Simulation Parameter & Value \\
    \midrule
    Sim $dt$ & 1/60~s \\
    Control $dt$ & 1/30~s \\
    Number of envs & 1024 \\
    Episode length & 300 \\
    \midrule
    Number of substeps & 2 \\
    Solver & TGS \\
    Number of position iterations & 4 \\
    Number of velocity iterations & 0 \\
    Contact offset & 0.02 \\
    Rest offset & 0.0 \\
    Bounce threshold velocity & 0.2 \\
    Max depenetration velocity & 10.0 \\
    \midrule
    Ground friction & 1.0 \\
    Ground restitutions & 0.0 \\
    \midrule
    Object density & 200 \\
    Object max convex hulls & 64 \\
    \bottomrule
  \end{tabular}
  }
\end{table}

\begin{table}[t]
  \centering
  \caption{
  Range of Motion of the humanoid robot used in this paper. \emph{Body} denotes body joints—hip, knee, ankle, toe, torso, spine, chest, neck, head, thorax, shoulder, elbow, and wrist.}
  \label{tab:supp_rom}
  \scalebox{0.85}{
  \begin{tabular}{l|cc cc}
    \toprule
    \multirow{2}{*}{Joint} & \multicolumn{2}{c}{$x$-axis} & \multicolumn{2}{c}{$y$- and $z$-axis} \\
    \cmidrule(lr){2-3}\cmidrule(lr){4-5}
    & min & max & min & max \\
    \midrule
    Body & -180.000 & 180.000 & -180.000 & 180.000 \\
    Thumb1 & -55.625 & 55.625 & -55.625 & 55.625 \\
    Thumb2 & -5.625 & 5.625 & -5.625 & 5.625 \\
    Thumb3 & -5.625 & 90.000 & -5.625 & 5.625 \\
    Index1,2 & -55.625 & 55.625 & -5.625 & 5.625 \\
    Index3 & -5.625 & 90.000 & -5.625 & 5.625 \\
    Middle1,2 & -55.625 & 55.625 & -5.625 & 5.625 \\
    Middle3 & -5.625 & 90.000 & -5.625 & 5.625 \\
    Ring1,2 & -55.625 & 55.625 & -5.625 & 5.625 \\
    Ring3 & -5.625 & 90.000 & -5.625 & 5.625 \\
    Pinky1,2 & -55.625 & 55.625 & -5.625 & 5.625 \\
    Pinky3 & -5.625 & 90.000 & -5.625 & 5.625 \\
    \bottomrule
  \end{tabular}
  }
\end{table}

\section{Motion Styles and Interaction Categories}
\label{sec:supp_task}

We evaluate our method against baselines across four dynamic motion styles and five interaction categories to assess its scalability to diverse motions and object types.

\subsection{Dynamic Motion Styles}
\label{subsec:supp_motion}

The following outlines the expected agent behavior and success criteria for each dynamic motion style. For $SR$ evaluation, an episode is considered success if the agent meets the style-specific criteria and keeps a target object within $0.3$ m of the reference path without dropping it.

\begin{itemize}
\item \textbf{\emph{Run Forward.}} \\
Success is achieved if the agent's horizontal pelvis velocity reaches within $0.5$~m/s of the reference's maximum horizontal pelvis velocity:
\[
\max_n \dot{p}_{\text{pelvis,xy}}[n] \ge \max_n \dot{\bar{p}}_{\text{pelvis,xy}}[n] - 0.5,
\]
where $\dot{p}_{\text{pelvis,xy}}[n]$ and $\dot{\bar{p}}_{\text{pelvis,xy}}[n]$ denote the horizontal pelvis velocity at timestep $n$ of the rollout and the reference, respectively.

\vspace{3pt}

\item \textbf{\emph{Jump Forward.}} \\
Succeed if the pelvis height gain reaches at least $50\%$ of the reference max pelvis height and the number of false foot-floor contacts remains below $10$.
A false foot-floor contact occurs when a foot touches the floor in the rollout but not in the reference at time step $n$. 
\begin{gather*}
\frac{\max_n{(p_\text{pelvis,z}[n])} - p_\text{pelvis,z}^\text{T}}{\max_n{(\bar{p}_\text{pelvis,z}[n])} - p_\text{pelvis,z}^\text{T}} \ge 0.5, \\
\sum_{n=0}^{N-1} c_\text{foot}[n] (1 - \bar{c}_\text{left foot}[n]) \le 10,
\end{gather*}
where $p_\text{pelvis,z}[n]$ and $\bar{p}_\text{pelvis,z}[n]$ denote the pelvis height of the rollout and the reference, respectively. $p_\text{pelvis,z}^\text{T}$ denotes the initial pelvis height when the agent is standing with a T-pose. $c_\text{foot}[n]$ and $\bar{c}_\text{foot}[n]$ are binary foot-floor contact states for the rollout and the reference, respectively. 

\vspace{3pt}

\item \textbf{\emph{High Kick.}} \\
Succeed if the foot height reaches within $5$~cm of the reference maximum foot height. 
\[
\max_n{p_\text{foot,z}[n]} \ge \max_n{\bar{p}_\text{foot,z}[n]} - 0.05.
\]

\vspace{3pt}

\item \textbf{\emph{Dance.}} \\
An episode is considered success if pelvis joint height does not fall below $0.3$~m.
\[
\min_n{p_\text{pelvis,z}[n]} \ge 0.3.
\]
where $p_\text{pelvis,z}[n]$ denote the pelvis height of rollout.
\end{itemize}

\subsection{Interaction Categories}
\label{subsec:supp_category}

\begin{figure}[t]
  \centering
\includegraphics[width=\columnwidth]{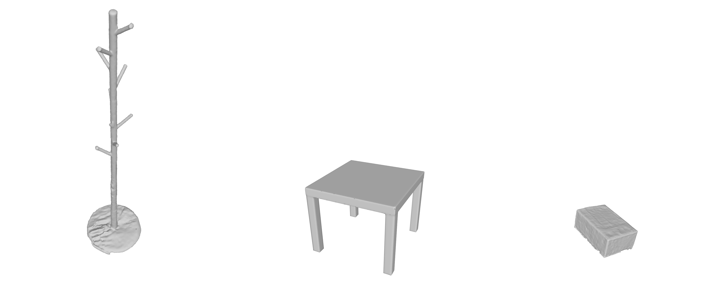}
  \caption{
  Visualization of \emph{clothesstand}, \emph{largetable}, and \emph{smallbox} from FullBodyManip~\cite{li2023object}, each preprocessed into 64 convex hulls following InterMimic~\cite{xu2025intermimic}.}
  \label{fig:supp_objects}
  \vspace{-10pt}
\end{figure}

We instantiate the above motion styles on three representative objects—\emph{smallbox}, \emph{clothesstand}, and \emph{largetable}—from the FullBodyManip~\cite{li2023object} dataset, as shown in Fig.~\ref{fig:supp_objects}. These objects represent a range of manipulation difficulty: \emph{smallbox} requires an easy two-hand manipulation, \emph{clothesstand} involves a one-hand interaction, and \emph{largetable} represents a challenging two-hand manipulation with a large object.

We evaluate five interaction categories using the three objects:
(i) \textbf{smallbox} involves lifting a smallbox with two hands; (ii) \textbf{clothesstand left} and (iii) \textbf{clothesstand right} involve lifting a clothesstand with the left and right hand, respectively; (iv) \textbf{largetable carry} denotes carrying a largetable at pelvis height using two hands; and (v) \textbf{largetable lift} represents lifting a largetable over the head with two hands.
Left- and right-hand \emph{clothesstand} interactions are treated as distinct tasks due to their asymmetric motion dynamics.

\subsection{Evaluation Metrics}
\label{subsec:supp_metrics}

We assess text-to-dynamic HOI generation with respect to HOI quality~\cite{li2023object}, physical plausibility~\cite{yuan2023physdiff}, and motion quality~\cite{tevet2023human}. To compare with DAViD~\cite{david}, we evaluate only motion segments starting at the onset of object contact.

\begin{itemize}
    \item \textbf{HOI quality.}
    \begin{itemize}
        \item \textbf{$C_\%$ $\uparrow$ (Contact Percentage).}  
        Percentage of frames in which any body part is in contact with the object.
        \item \textbf{$C_\text{cons}$ $\downarrow$ (Contact Consistency).}
        Standard deviation of object-vertex positions expressed in the hand’s local coordinate frame.
        A lower value indicates more consistent relative hand–object pose, reflecting more stable interaction.
    \end{itemize}
    \item \textbf{Physical plausibility.}
    \begin{itemize}
        \item \textbf{$\text{Pene}_\text{obj}$ $\downarrow$ (Object-floor Penetration, cm).}
        Measures interpenetration depth between the object and the floor, indicating violations of physical plausibility.
        \item \textbf{Skate $\downarrow$ (mm).}  
        Mean horizontal displacement of foot–ground contact points during contact.
        \item \textbf{Float $\downarrow$ (mm).}  
        Minimum distance between the ground and the lowest body joint.
        \item \textbf{$\text{Jitter}_\text{pos}$ $\downarrow$ ($\mathrm{m}/\mathrm{s}^3$).}
        Mean third-order derivative of joint positions, measuring motion smoothness.
    \end{itemize}
    \item \textbf{Human motion quality.}
    \begin{itemize}
        \item \textbf{R-Precision $\uparrow$.}  
        Retrieval-based relevance metric checking whether the ground-truth text–motion pair appears within the top-3 retrieved candidates.
        \item \textbf{Diversity $\uparrow$.}  
        Measures variability across the generated human motions.
    \end{itemize}
\end{itemize}

We evaluate dynamic HOI imitation using the metrics introduced in InterMimic~\cite{xu2025intermimic}, along with the training time.

\begin{itemize}
    \item \textbf{$SR \uparrow$ (Success Rate).} 
    Fraction of episodes that satisfy the success criterion in Sec.~\ref{subsec:supp_motion} without dropping the object.
    \item \textbf{$T$ (Training Time, h).}
    Training time until the learning curve converges.
    \item \textbf{$D \uparrow$ (Mean Duration, $\mathrm{s}$).} 
    Average uninterrupted interaction time per episode without fall or drop.
    \item \textbf{$E_\text{HOI}$ $\downarrow$}
    Mean per-frame Euclidean distance between the planned and executed HOI motions across 21 SMPL joints (excluding the pelvis) and the object.  
    The metric is computed after globally aligning the pelvis joint at $n{=}0$.
    \item \textbf{$\text{Jitter}_\text{DoF}$ $\downarrow$ ($\textrm{rad}/\textrm{s}^3$)}
    Third-order derivative of joint rotations, measuring motion smoothness in the joint-rotation space.  
    Each DoF corresponds to a rotational axis of a joint.
\end{itemize}

\section{Additional Implementation Details}
\label{sec:supp_impl_details}

\begin{table}[t]
  \centering
  \caption{
  Hyperparameters for two imitation experts and our composer training.}
  \label{tab:supp_hyperparams}
  \scalebox{0.85}{
  \begin{tabular}{ll|c}
    \toprule
    & Hyperparameter & Value \\
    \midrule
    PHC & input size & 574 \\
    & action size & 69 \\
    & number of primitives & 3 \\
    & \multirow{2}{*}{architecture (primitive)} & [2048, 1536, 1024, \\
    && 1024, 512, 512] \\
    & \multirow{2}{*}{architecture (composer)} & [2048, 1536, 1024, \\
    && 1024, 512, 512] \\
    \midrule
    InterMimic & input size & 3198 \\
    & action size & 153 \\
    & architecture & [1024, 1024, 512] \\
    \midrule
    Composer & input size & 3630 \\
    & action size ($w$) & 153 \\
    & action size ($r$) & 153 \\
    & action size ($\mu$) & 4 \\
    & \multirow{2}{*}{architecture} & [1024, 1024, \\
    && 512, 512] \\
    & PCA buffer size $B$ & 16 \\
    & PCA subspace dim $S$ & 4 \\
    & Extrapolation coefficient $\rho$ & 0.2 \\
    & Subspace coefficient $\sigma$ & 0.08 \\
    \midrule
    Training & Discount Factor & 0.99 \\
    & Generalized adv. estimation & 0.95 \\
    & Entropy reg. coefficient & 0.0 \\
    & Optimizer & Adam \\
    & Actor learning rate & 2e-5 \\
    \vspace{-2pt}
    & Actor learning rate  & \multirow{2}{*}{2e-6} \\
    &(for $\mathrm{InterMimic}_\mathrm{FT}$)\\
    & Critic learning rate & 1e-4 \\
    & Minibatch size & 16384 \\
    & Horizon length $H$ & 32 \\
    & Max episode length $N$ & 300 \\
    \bottomrule
  \end{tabular}
  }
\end{table}

The interaction onset is defined as 1.5 seconds after the initial contact, and we observed that the method is robust to this choice. After the onset, the local rotations of interaction-related joints $\mathcal{J}_{\text{int}}$ — thorax, shoulder, elbow, and wrist — are fixed to maintain consistent contacts.

All pretrained imitation experts~\cite{Luo2023PerpetualHC,xu2025intermimic} used in this paper follow their publicly available implementations. InterMimic serves as our full-body controller that predicts PD targets for the entire body joints including the fingers. It receives both the observation $\textbf{s}[n]$ and reference states $\textbf{g}[n]$ as input and uses a 3-layer MLP followed by a final linear layer producing PD actions. 
PHC is a versatile expert specialized for high-dynamics whole-body imitation, producing PD targets for all body joints except the fingers. We use the PHC+ variant with three primitive agents, each implemented as a 6-layer MLP trained on progressively more difficult subsets, and a 6-layer composer network that blends their outputs. We follow the settings of the original paper. All InterMimic and PHC components are kept frozen during our composer training.

We implement several baselines for comparison. PPO denotes a policy trained entirely from scratch using the same observation and action spaces as our method. $\textrm{PHC}_\textrm{R}$ and $\textrm{InterMimic}_\textrm{R}$ attach an additional residual MLP that receives the same $\textbf{s}[n]$ and $\textbf{g}[n]$ as inputs and outputs an additive correction term; residual outputs are directly added to those of the pretrained policy. $\textrm{InterMimic}_\textrm{FT}$ corresponds to training the pretrained InterMimic model without freezing any parameters. We exclude $\textrm{PHC}_\textrm{FT}$ since PHC does not receive object observations.

Our proposed composer network is a lightweight MLP designed to blend the two experts spatio-temporally. 
The composer network is a 4-layer MLP followed by a final linear layer that predicts a 310-dimensional weight vector. The first 153 dimensions correspond to the per-DoF interpolation weights $\mathbf{w}[n]$, the next 153 dimensions to the bounded extrapolation weights $\mathbf{r}[n]$, and the remaining 4 dimensions to the exploration coefficients $\boldsymbol{\mu}[n]$. These components are then passed through separate nonlinear heads: $\mathrm{Sigmoid}(\cdot)$ for $\mathbf{w}[n]$, $\rho \tanh(\cdot)$ for $\mathbf{r}[n]$, and $\sigma \tanh(\cdot)$ for $\boldsymbol{\mu}[n]$.
For PCA-based low-dimensional action subspace exploration, at each timestep we store the body-only action difference between the two experts, $\Delta \textbf{a}_\textrm{body}[n] = \textbf{a}_\textrm{body}^\textrm{IM}[n] - \textbf{a}^\textrm{PHC}[n]$, into a buffer. PCA is performed on the most recent $B$ samples, and the top-$S$ eigenvectors form an orthogonal basis for the exploration subspace. The detailed hyperparameters are summarized in Tab.~\ref{tab:supp_hyperparams}.


\begin{figure}[t]
  \centering

  \begin{minipage}[c]{0.54\columnwidth}
    \captionsetup{type=figure}
    \centering    
    \includegraphics[width=0.9\linewidth]{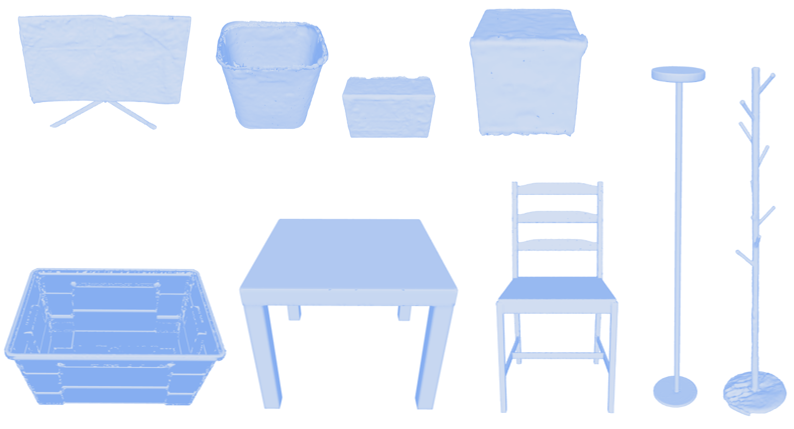}
    \caption{FullBodyManip objects used in $\text{Ours}_+$.}
    \label{fig:objects}
  \end{minipage}
  \hfill
  \begin{minipage}[c]{0.44\columnwidth}
    \captionsetup{type=figure}
    \centering
    \vspace{2pt}

    \begin{subfigure}[t]{0.48\linewidth}
      \centering
      \includegraphics[width=\linewidth]{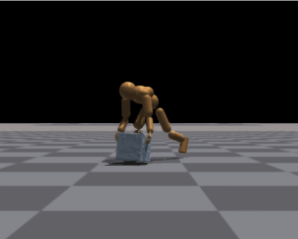}
      \captionsetup{skip=1pt}
      \caption{Push}
      \label{fig:exec_push}
    \end{subfigure}
    \hfill
    \begin{subfigure}[t]{0.48\linewidth}
      \centering
      \includegraphics[width=\linewidth]{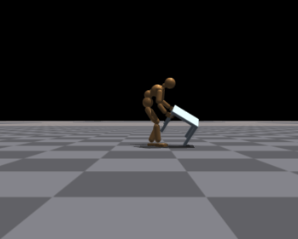}
      \captionsetup{skip=1pt}
      \caption{Drag}
      \label{fig:exec_drag}
    \end{subfigure}

    \vspace{-2pt}

    \begin{subfigure}[t]{0.48\linewidth}
      \centering
      \includegraphics[width=\linewidth]{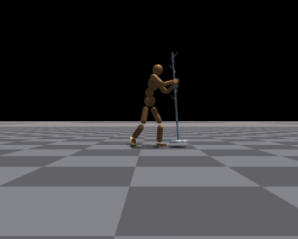}
      \captionsetup{skip=1pt}
      \caption{Swing}
      \label{fig:exec_swing}
    \end{subfigure}
    \hfill
    \begin{subfigure}[t]{0.48\linewidth}
      \centering
      \includegraphics[width=\linewidth]{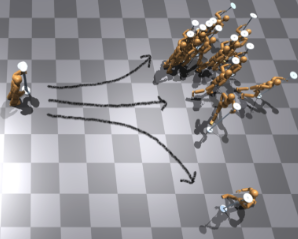}
      \captionsetup{skip=1pt}
      \caption{Plan path}
      \label{fig:exec_div}
    \end{subfigure}

    \vspace{-6pt}
    \caption{Additional motion styles.}
    \label{fig:exec_half}
  \end{minipage}

\end{figure}

\begin{table*}[t]
\centering
\caption{Additional motion quality results. For clarity, we repeat the main quantitative results of planning-only methods.}
\label{tab:supp_motion_quality}
\setlength{\tabcolsep}{4pt}
\scalebox{0.85}{
\begin{tabular}{lcccccccccc}
\toprule
\multirow{2}{*}{Method} & \multirow{2}{*}{Phys.}
& \multicolumn{2}{c}{HOI quality}
& \multicolumn{4}{c}{Physical plausibility}
& \multicolumn{2}{c}{Motion quality} \\
\cmidrule(lr){3-4} \cmidrule(lr){5-8} \cmidrule(lr){9-10}
& 
& $C\% \uparrow$ & $C_{\mathrm{cons}} \downarrow$
& Pene. $\downarrow$ & Skate $\downarrow$ & Float $\downarrow$ & Jitter$_{\mathrm{pos}} \downarrow$
& R-Prec $\uparrow$ & Div. $\uparrow$ \\
\midrule

\multicolumn{10}{l}{\textbf{Planning-only}} \\
DAViD      &            & 0.848 & 10.9  & 4.842 & 0.261 & 20.4 & $6.69 \times 10^4$ & 0.310 & 6.70 \\
HOI-Diff   &            & 0.358 & 22.8  & 2.785 & 0.633 & 17.8 & $1.56 \times 10^4$ & 0.257 & 4.66 \\
Ours$_P$   &            & \textbf{1.000} & \textbf{0.906} & 4.196 & \textbf{0.217} & 30.1 & $5.93 \times 10^4$ & \textbf{0.332} & 5.56 \\
\midrule

\multicolumn{10}{l}{\textbf{Planning+Execution}} \\
InterMimic$_{\mathrm{FT}}$ & \checkmark & 0.999 & 2.50 & {0.000} & 2.66 & {7.98} & $7.05 \times 10^4$ & 0.256 & 2.99 \\
InterMimic$_R$             & \checkmark & 0.999 & 3.07 & {0.000} & {0.512} & 19.6 & $5.43 \times 10^4$ & 0.216 & 3.40 \\
Ours$_{P+E}$               & \checkmark & \textbf{0.999} & {2.95} & {0.009} & {0.786} & {9.95} & $\boldsymbol{4.53 \times 10^4}$ & \textbf{0.316} & \textbf{3.97} \\
\bottomrule
\end{tabular}
}
\end{table*}

\begin{table}[t]
\centering
\caption{Additional ablation studies. The final row denotes $\text{Ours}_+$ which is our method trained and evaluated on a larger set of objects shown in Fig.~\ref{fig:objects}. }
\label{tab:supp_ablation}
\resizebox{\columnwidth}{!}{
\begin{tabular}{llcccc}
\toprule
\multicolumn{2}{c}{Method} & \multirow{2}{*}{More Obj.} & \multirow{2}{*}{$SR \uparrow$} & \multirow{2}{*}{$D \uparrow$} & \multirow{2}{*}{$E_{\mathrm{HOI}} \downarrow$} \\
\cmidrule(lr){1-2}
\multicolumn{1}{l}{Planning} & \multicolumn{1}{l}{Execution} & & & & \\
\midrule
DAViD    & \multirow{2}{*}{$\text{Ours}_\text{MLP+PCA}$} &  & --    & 0.555 & 13.47 \\
HOI-Diff &                       &  & --    & 0.437 & 23.09 \\
\midrule
\multirow{4}{*}{$\text{Ours}_\text{P}$} & Heuristic$_{\text{Hand}}$    &  & 0.365 & 1.069 & 13.04 \\
                       & Heuristic$_{\text{Arm}}$     &  & 0.472 & 1.603 & 17.18 \\
                       & Hard MoE         &  & 0.383 & 2.748 & 20.48 \\
                       & Hard MoE (Joint) &  & 0.383 & 2.535 & 26.45 \\
\midrule
$\text{Ours}_\text{P}$ & $\text{Ours}_\text{MLP+PCA}$ &  & \textbf{0.591} & \textbf{4.607} & \textbf{11.67} \\
$\text{Ours}_\text{P}$ & $\text{Ours}_\text{MLP+PCA}$ & \checkmark & 0.453 & 4.348 & 11.99 \\
\bottomrule
\end{tabular}
}
\end{table}

\section{Additional Qualitative Results}
\label{sec:supp_more_qual_results}
\begin{figure}[t]
    \centering
    \includegraphics[width=0.9\columnwidth]{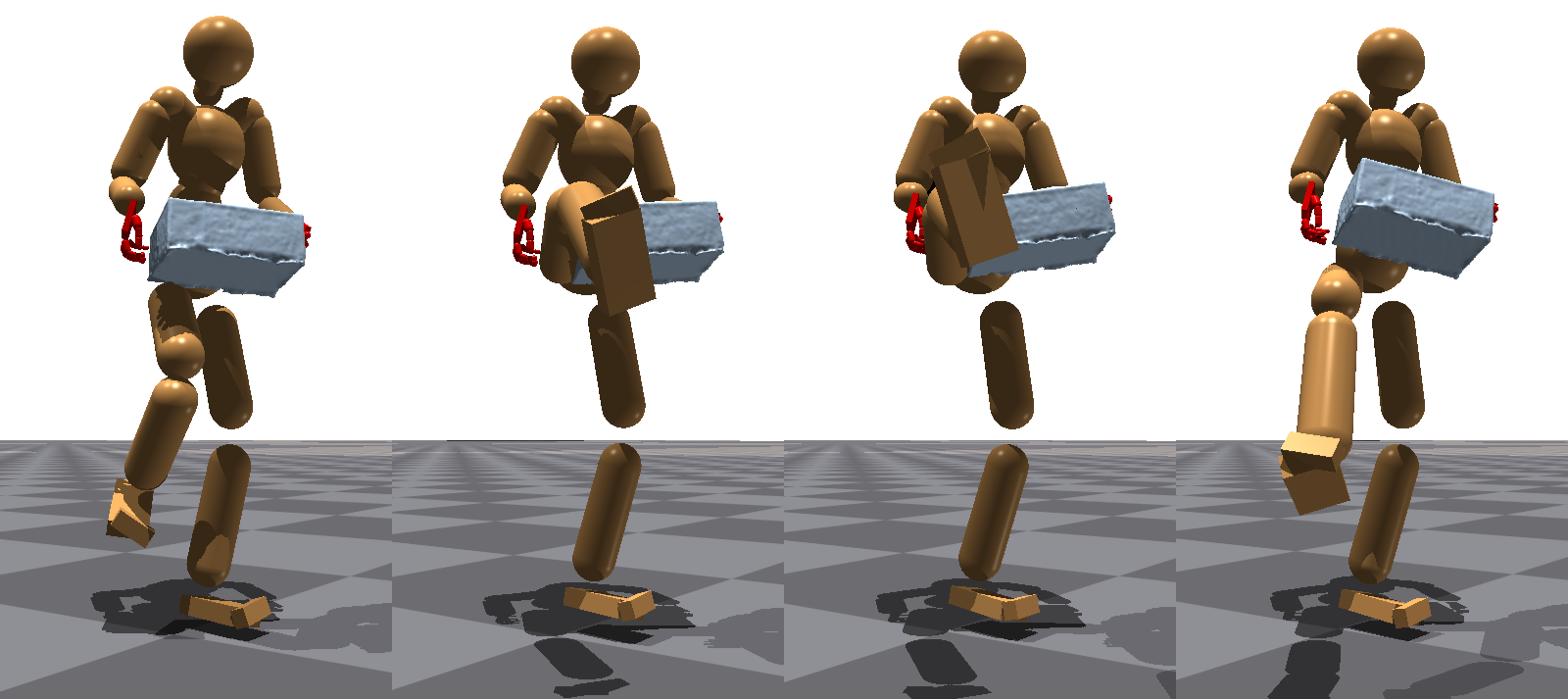}
    \caption{
    The HOI motion planner does not enforce explicit physical constraints, causing non-physical poses such as body–object penetration and unstable hand object contacts.}
    \label{fig:supp_failure_planning}
\end{figure}

\begin{figure*}[t]
\centering

\begin{minipage}[b]{0.10\textwidth}
    \small \raggedleft MDM
\end{minipage}
\begin{minipage}[b]{0.4\textwidth}
    \includegraphics[width=\linewidth]{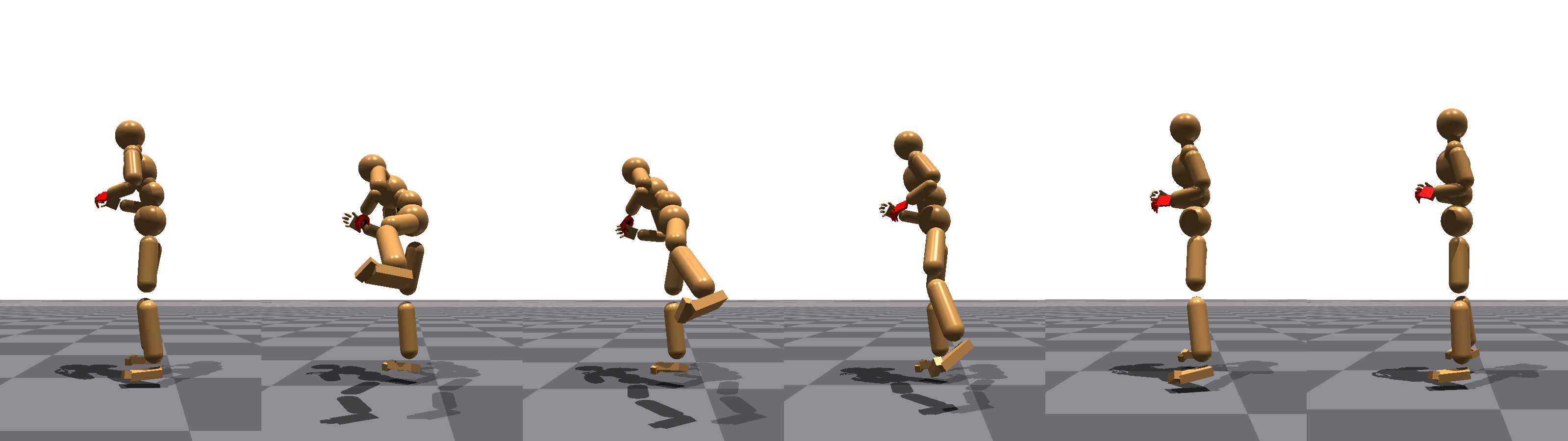}
\end{minipage}
\begin{minipage}[b]{0.4\textwidth}
    \includegraphics[width=\linewidth]{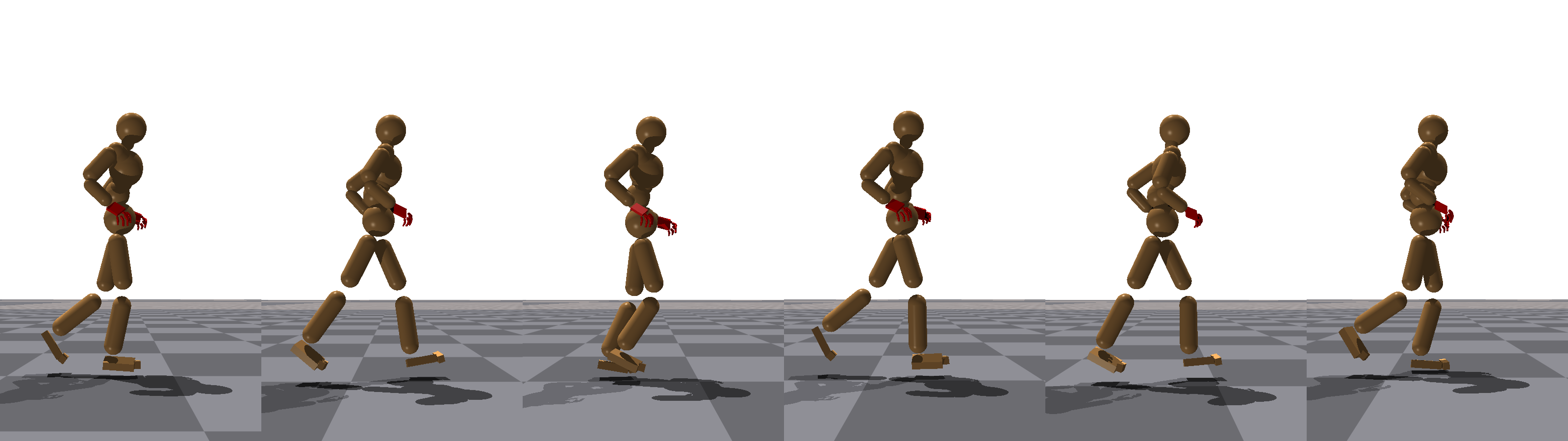}
\end{minipage}

\begin{minipage}[b]{0.10\textwidth}
    \small \raggedleft DAViD
\end{minipage}
\begin{minipage}[b]{0.4\textwidth}
    \includegraphics[width=\linewidth]{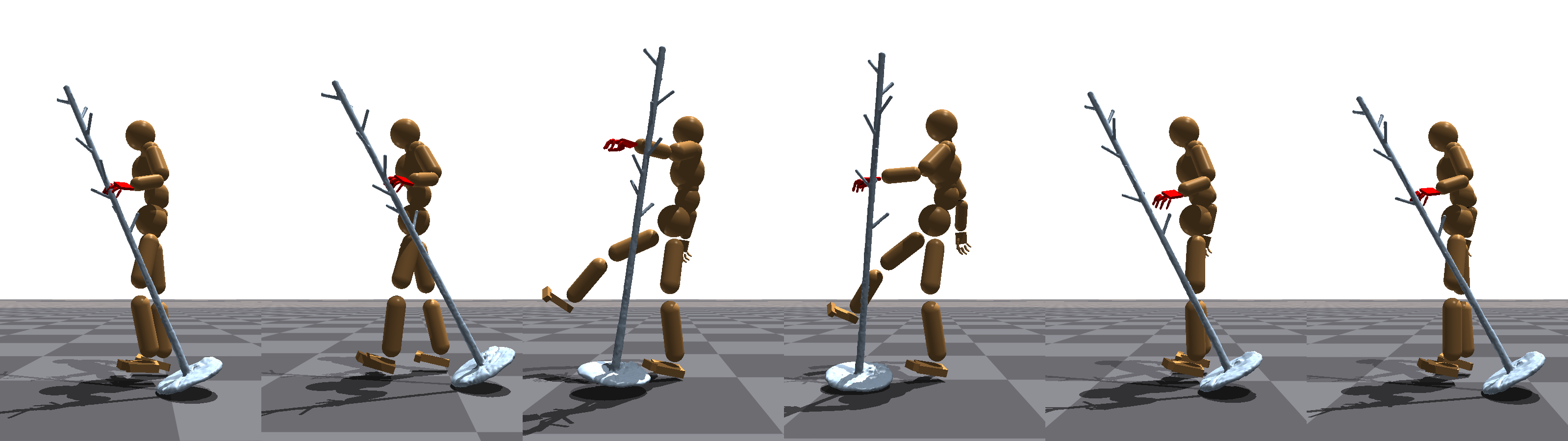}
\end{minipage}
\begin{minipage}[b]{0.4\textwidth}
    \includegraphics[width=\linewidth]{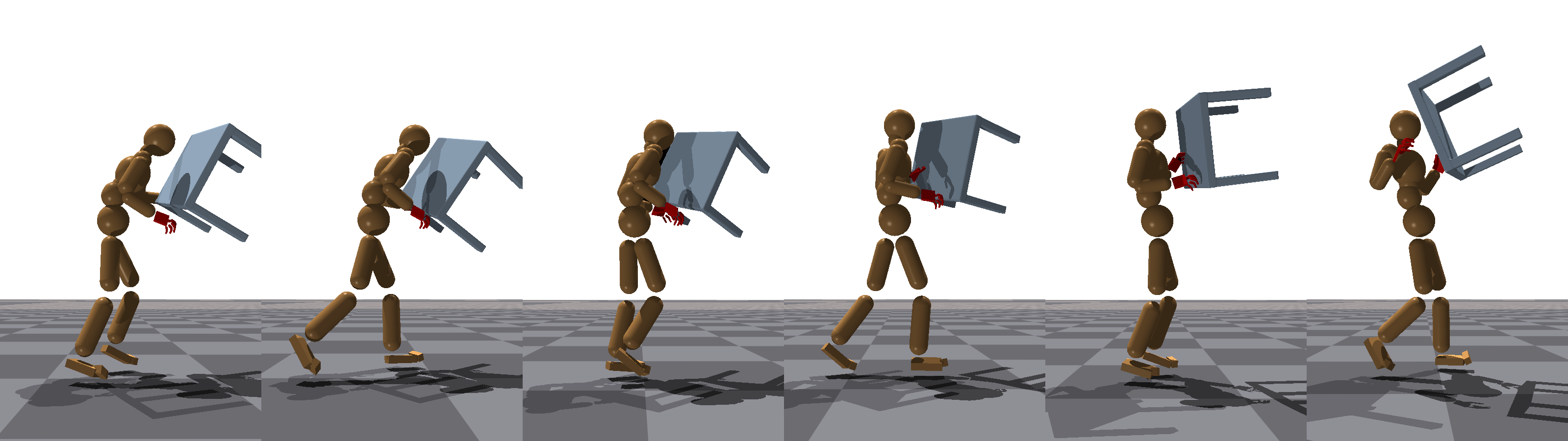}
\end{minipage}

\vspace{6pt}

\begin{minipage}[b]{0.10\textwidth}
    \small \raggedleft Ours
\end{minipage}
\begin{minipage}[b]{0.4\textwidth}
    \includegraphics[width=\linewidth]{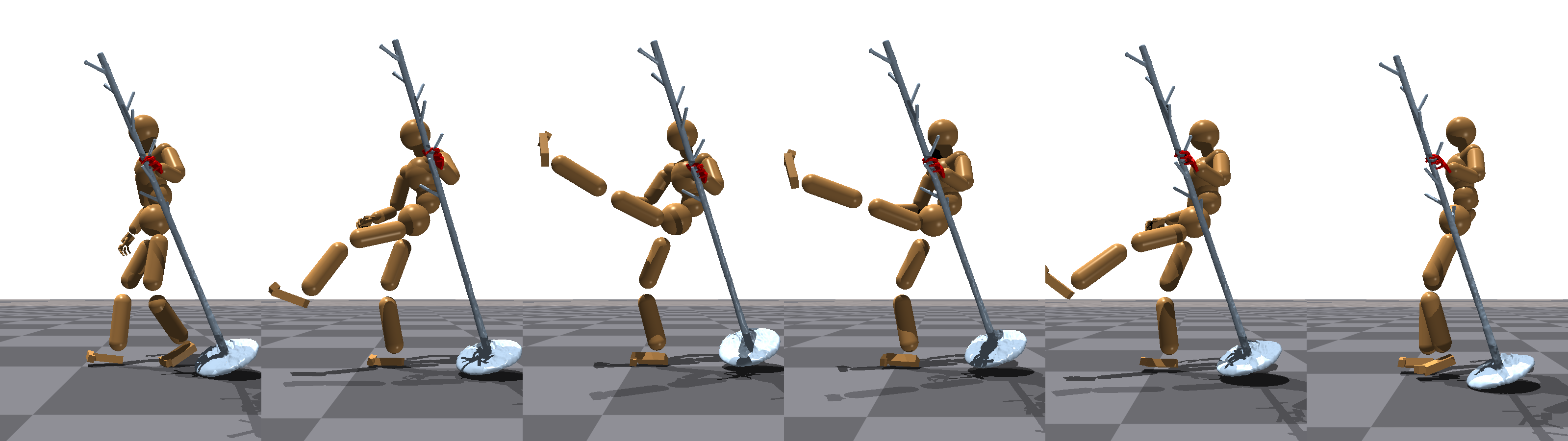}
\end{minipage}
\begin{minipage}[b]{0.4\textwidth}
    \includegraphics[width=\linewidth]{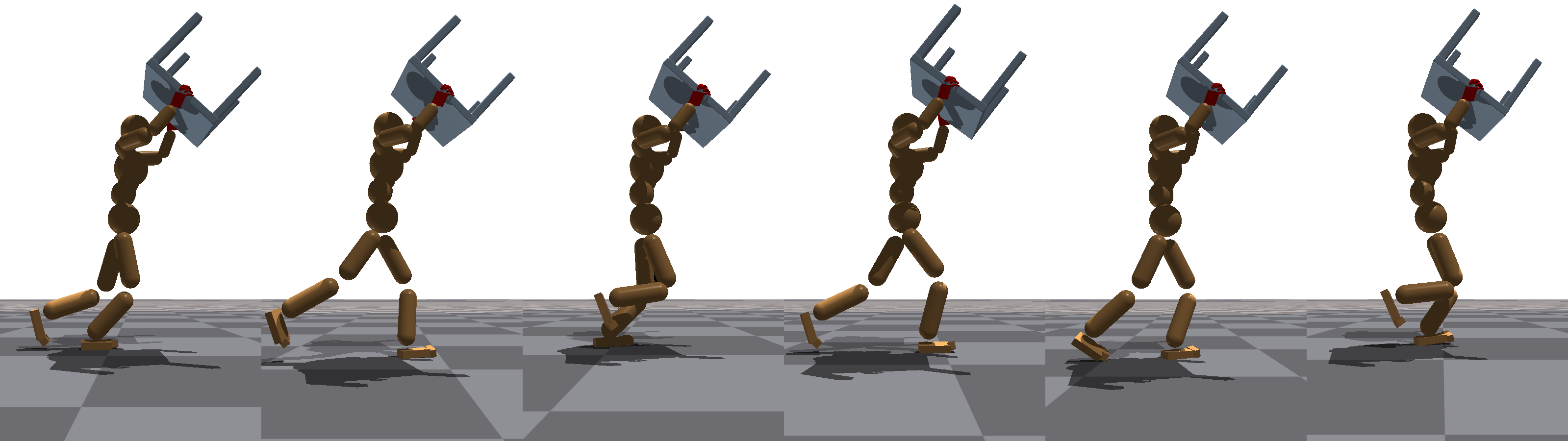}
\end{minipage}

\begin{minipage}[b]{0.10\textwidth}
    \rule{0pt}{1ex}
\end{minipage}
\begin{minipage}[b]{0.4\textwidth}
    \small \centering A person kicks, clothesstand
\end{minipage}
\begin{minipage}[b]{0.4\textwidth}
    \small \centering A person runs, largetable
\end{minipage}

\caption{
\textbf{Qualitative comparisons of prior-blending for HOI planning.}
MDM~\cite{tevet2023human} does not account for object interaction, often failing to establish hand–object contacts.
DAViD~\cite{david} produces unstable and inconsistent hand–object alignment.
In contrast, our planning stage maintains consistent hand–object interaction while preserving the intended motion style.
}
\label{fig:supp_qualitative_planning}
\end{figure*}

\begin{figure*}[t]
  \centering
  \includegraphics[width=\textwidth]{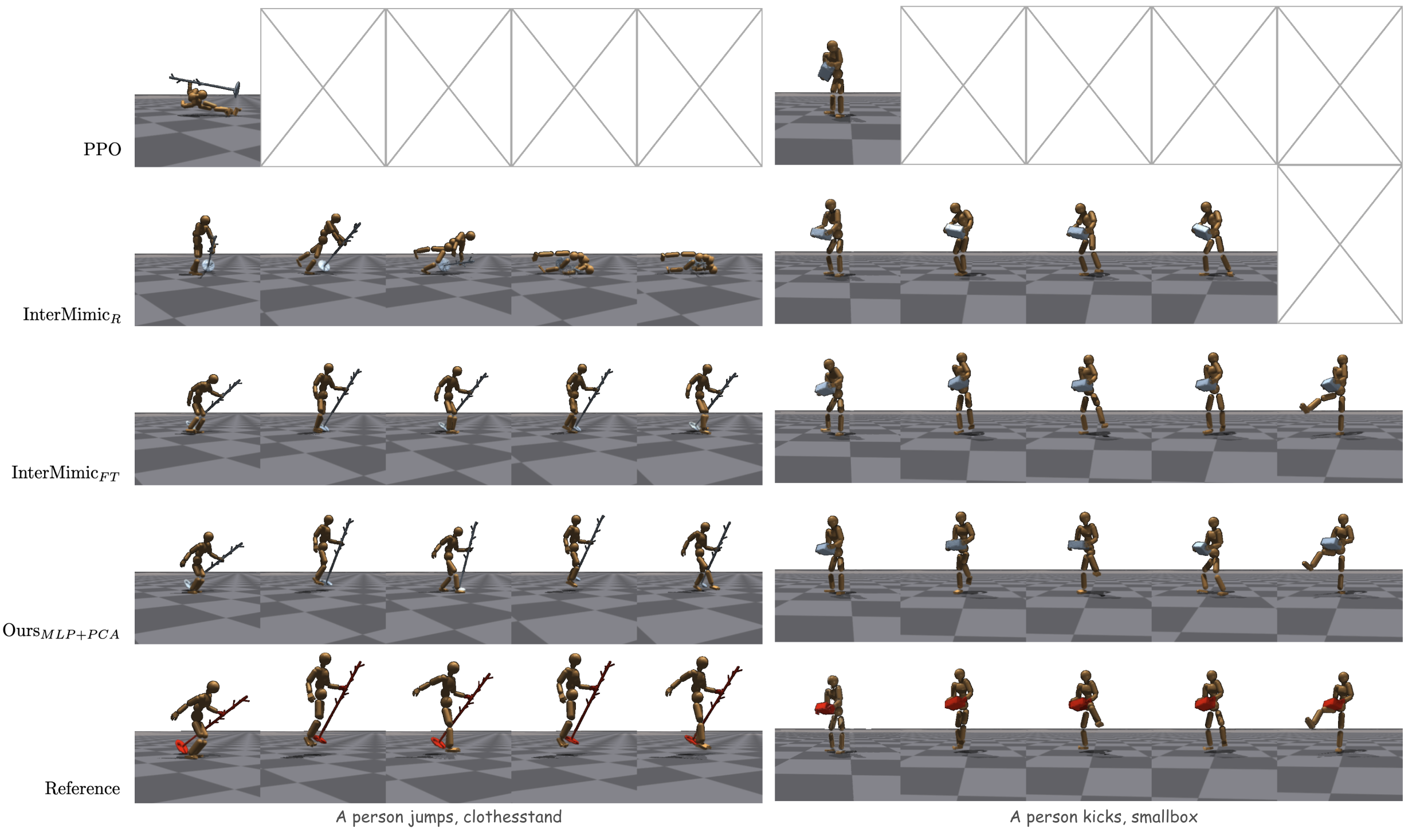}
  \caption{\textbf{Qualitative comparisons of HOI execution.}
  Our method generates more physically plausible and dynamic motions than baseline controllers.
  During jumps, $\textrm{InterMimic}_\textrm{FT}$ approximates the motion through small repetitive stepping patterns, whereas $\textrm{Ours}$ produces an actual two-foot lift-off.
  During kicks, our foot trajectories reach higher and align more closely with the reference.
  Frames marked with \textbf{X} indicate early termination due to object drops or robot falls.
  }
  \label{fig:supp_qualitative_stage2}
\end{figure*}

Prior-blending planning comparisons are shown in Fig.~\ref{fig:supp_qualitative_planning}.
The execution-stage comparisons of our composer-based imitation against baselines are presented in Fig.~\ref{fig:supp_qualitative_stage2}.
We compare $\text{PPO}$, $\text{InterMimic}_\textrm{R}$, $\text{InterMimic}_\textrm{FT}$, and $\text{Ours}_{\text{MLP+PCA}}$ against the reference motions produced by our planning stage, evaluated across additional object categories and dynamic motion styles. Our method achieves the highest imitation performance across object types and motion styles.

\paragraph{Failure Cases.}
At the planning stage, the lack of explicit physical constraints leads to non-physical poses such as object–body penetration or unstable contacts, as illustrated in Fig.~\ref{fig:supp_failure_planning}.

\section{Additional Ablation Studies}
\label{sec:supp_more_ablations}

\subsection{Prior-Blending for Dynamic HOI Planning}
\label{subsec:ablation_stage1}
Tab.~\ref{tab:supp_motion_quality} (rows 1--3) compares our Prior blending $\text{Ours}_\text{P}$ with HOI motion planners, DAViD and HOI-Diff.
We measure motion-dynamic style text alignment by R-Prec and contact stability by $C_{\%}$ and $C_{\text{cons}}$.
$\text{Ours}_\text{P}$ achieves the best for the metrics, indicating stable contacts while generating dynamic motions.
Tab.~\ref{tab:supp_ablation} (rows 1--2,7) measures performance of our composer, using HOI plans generated by each method; ours performs the best, showing that Prior-blending is essential for composer training.

\subsection{Composer-based Dynamic HOI Execution}
\label{subsec:ablation_stage2}
Tab.~\ref{tab:supp_ablation} (rows 3--7) compares our composer with alternative blending baselines: Heuristic$_{\text{Hand}}$ (hands from InterMimic, others from PHC), Heuristic$_{\text{Arm}}$ (hands+arms from InterMimic, others from PHC), Hard MoE (global InterMimic/PHC switching), and Hard MoE (Joint) (per-joint switching).
Our composer-based blending achieves the highest SR, while Hard MoE (Joint) is even worse than Hard MoE despite finer control, suggesting that soft blending is more important than finer discrete switching.
Tab.~\ref{tab:supp_motion_quality} (rows 4--6) reports quality of the executed results for the next best baselines in the main paper, InterMimic$_{\text{FT}}$ and InterMimic$_{\text{R}}$; Ours achieves higher R-Prec, that shows how the method preserves the dynamics of reference motion.

\section{Discussions}
\label{sec:supp_discussions}

\subsection{related works}
OmniGrasp~\cite{luo2024omnigrasp} focuses primarily on trajectory-conditioned control, yet supports neither text-guided nor dynamic HOI. TokenHSI~\cite{pan2025tokenhsi} learns primitive skills such as walking, carrying a box, and climbing on a box, then composing them using a transformer architecture for complex tasks. However, reliance on primitive skills for specific objects (e.g., box) and motion style limits its scalability to novel and dynamic HOI tasks.

\subsection{Scalability}
Tab.~\ref{tab:supp_ablation} (rows 7--8) reports $\text{Ours}_{\text{+}}$, where we expand object categories with different geometries and weights shown in Fig.~\ref{fig:objects}.
$\text{Ours}_{\text{+}}$ achieves comparable performance to Ours, indicating geometry- and weight-agnostic execution of our blending strategies.
Fig.~\ref{fig:exec_push}--\ref{fig:exec_swing} further show that our approach generalizes to diverse interaction styles, push, drag, and swing.

\subsection{Plug-and-play Capability}
Each module in this work can be replaced in a plug-and-play manner.
Fig.~\ref{fig:exec_div} shows that replacing the prior-blending backbone with GMD~\cite{karunratanakul2023gmd} enables path planning.
Similarly, InterMimic can be replaced by dataset-specific HOI experts; e.g., a ParaHome~\cite{kim2024parahome}-trained agent can be plugged into the InterMimic branch for kitchen-centric interactions.

\subsection{Inpainting Strategy in Dynamic HOI Planning}
For efficient dynamic HOI agent training, reference motions should preserve coherent whole-body kinematics and contact consistency. Our prior-blending preserves kinematic coherence through iterative denoising~\cite{rombach2022high} and enforces contact consistency through inpainting. The resulting ``glued" artifacts are refined by the execution stage, yielding natural and physically-plausible grasping.

\subsection{Error Accumulation of Two-stage Framework}
CLoSD~\cite{tevet2025closd} tackles error accumulation between the planning and execution stages via an autoregressive closed loop. 
Instead, our work prioritizes dynamic HOI motion generation. 
Incorporating the autoregressive planner and our composer in closed-loop is a promising direction to mitigate error accumulation, which we leave for future work.

\subsection{Limitations}
While our method enables diverse and dynamic HOI generation and imitation, it has several limitations.
First, integrating the two experts used in this paper is challenging because they were originally trained under different humanoid kinematics, making it difficult to preserve their full capabilities.
Second, the planning stage does not enforce explicit physical constraints, such as preventing human–object penetration or dynamically infeasible human poses; therefore, physical plausibility is guaranteed only after the execution stage. 

\subsection{Future Work}
Future work includes expanding motion-style and interaction-category combinations to enable a broader range of dynamic HOI behaviors.
Furthermore, we plan to incorporate explicit physical feasibility constraints directly into the planning stage, reducing reliance on the execution stage for correction.
Additionally, our planning–execution framework can be extended in an autoregressive manner to enable real-time dynamic HOI manipulation.

\end{document}